\documentclass{article}

\PassOptionsToPackage{numbers, compress}{natbib}
\usepackage[pagebackref=true]{hyperref}

\usepackage[main, final]{neurips_2026}

\usepackage[utf8]{inputenc} 
\usepackage[T1]{fontenc}    
\usepackage{booktabs}       
\usepackage{amsfonts}       
\usepackage{nicefrac}       
\usepackage{microtype}      
\usepackage[table]{xcolor}  
\usepackage{pifont}         

\usepackage{graphbox}
\usepackage{multirow}
\usepackage{makecell}
\usepackage{wrapfig}
\usepackage{float}  
\usepackage[table]{xcolor}

\usepackage{caption}
\usepackage{float}
\floatname{algorithm}{Algorithm}
\newfloat{algorithm}{tbp}{loa}
\usepackage{algorithmic}
\usepackage{amsmath}
\usepackage{amssymb}
\usepackage{enumitem}
\usepackage{subcaption}
\usepackage{amsthm}
\newtheorem{theorem}{Theorem}[section]

\usepackage[capitalise]{cleveref}

\newcommand{\Comment}[1]{\small{\color{gray}\textit{$\triangleright$ #1}}}

\definecolor{mydarkgreen}{rgb}{0,0.45,0.08}
\definecolor{mydarkblue}{RGB}{30,129,176}
\definecolor{mydarkblue}{rgb}{0,0.1,0.6}

\hypersetup{ %
  colorlinks=true,
  linkcolor=mydarkblue,
  citecolor=mydarkblue,
  filecolor=mydarkgreen,
  urlcolor=mydarkgreen,
}

\title{SCALE: Improving Guidance-Free Visual Generation via Self-Contrastive Alignment for \\ Likelihood Estimation}
\author{
    \vspace{-25pt} \\
    \textbf{Jiwan Hur \quad\quad Dong-Jae Lee \quad\quad Dongyeun Lee \quad\quad Gyojin Han \quad\quad Junmo Kim} \\  
    KAIST, South Korea \\
    {\texttt{\small \{jiwan.hur, jhtwosun, ledoye, hangj0820, junmo.kim\}@kaist.ac.kr}} \vspace{3pt} \\
    Project page: \url{https://jiwanhur.github.io/SCALE} \vspace{-10pt}
}

\usepackage{amsmath,amsfonts,bm,amssymb,mathtools,amsthm}
\usepackage{color,xcolor,xspace}
\usepackage{booktabs}
\usepackage{thm-restate}

\def\eqref#1{Eq.~(\ref{#1})}

\def\1{\bm{1}}

\def\rvx{{\mathbf{x}}}

\def\vx{{\bm{x}}}

\DeclareMathAlphabet{\mathsfit}{\encodingdefault}{\sfdefault}{m}{sl}
\SetMathAlphabet{\mathsfit}{bold}{\encodingdefault}{\sfdefault}{bx}{n}

\def\gN{{\mathcal{N}}}

\DeclareMathOperator*{\argmax}{arg\,max}

\renewcommand{\emptyset}{\varnothing}

\begin{document}

\maketitle

\begin{abstract}
  Classifier-free guidance (CFG) is essential for improving sample quality in visual generation, but it incurs substantial sampling overhead. Existing guidance-free approaches halve this cost by approximating the CFG with a single model inference, but their performance is bounded by the CFG and inherits its limitations.
  In this paper, we propose \emph{self-contrastive guidance} (SCG), which contrasts the model's prediction on real and self-generated samples to obtain a direct, model-based signal that pushes generation toward the true data manifold.
  Integrating SCG into the standard likelihood objective via reparameterization yields \emph{Self-Contrastive Alignment for Likelihood Estimation} (SCALE), a guidance-free training framework that produces SCG-enhanced predictions in a single forward pass.
  Across class-conditional and text-to-image generation on diffusion and autoregressive models, SCALE consistently outperforms CFG-mimicking guidance-free baselines and improves further when combined with DPO, demonstrating that the self-contrastive signal yields gains complementary to both CFG and preference-based fine-tuning.
\end{abstract}

\section{Introduction}

Likelihood-based generative models, including diffusion~\cite{ho2020denoising,song2020score} and autoregressive (AR) models~\cite{vqvae2,llamagen}, have demonstrated remarkable scalability and performance.
Yet, for complex visual datasets, trained models often produce samples whose clarity falls short of the training images. Classifier-free guidance (CFG)~\cite{ho2022classifier} has thus become the de facto remedy.
By contrasting unconditional and conditional predictions, CFG effectively enhances fidelity, but at the cost of doubled sampling cost.

\begin{figure}[t]
  \centering
  \begin{subfigure}[b]{0.64\linewidth}
    {\includegraphics[width=\linewidth]{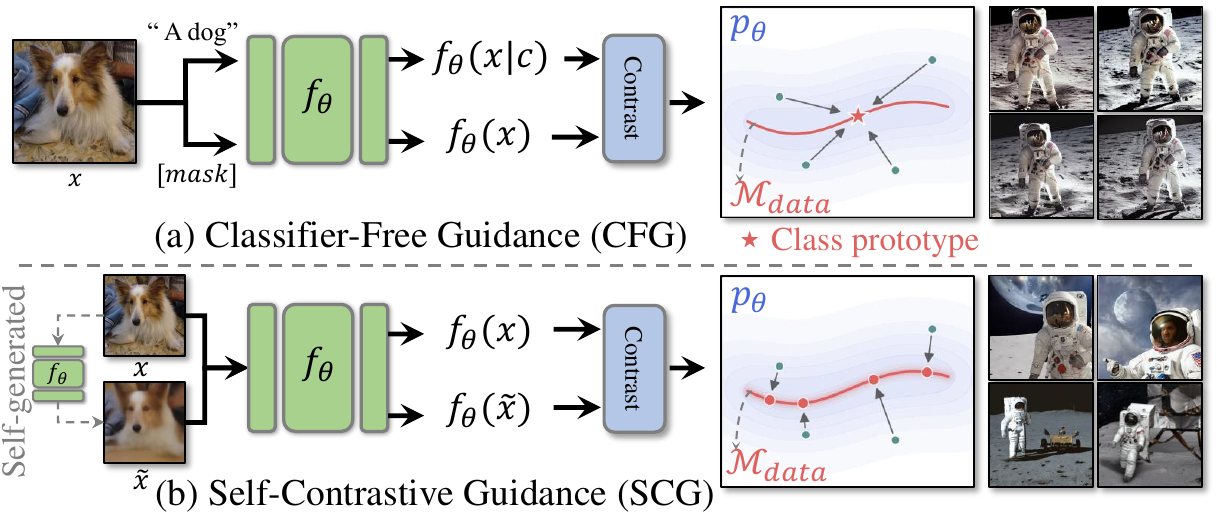}}
  \end{subfigure}\hfill
  \begin{subfigure}[b]{0.36\linewidth}
    \includegraphics[width=\linewidth]{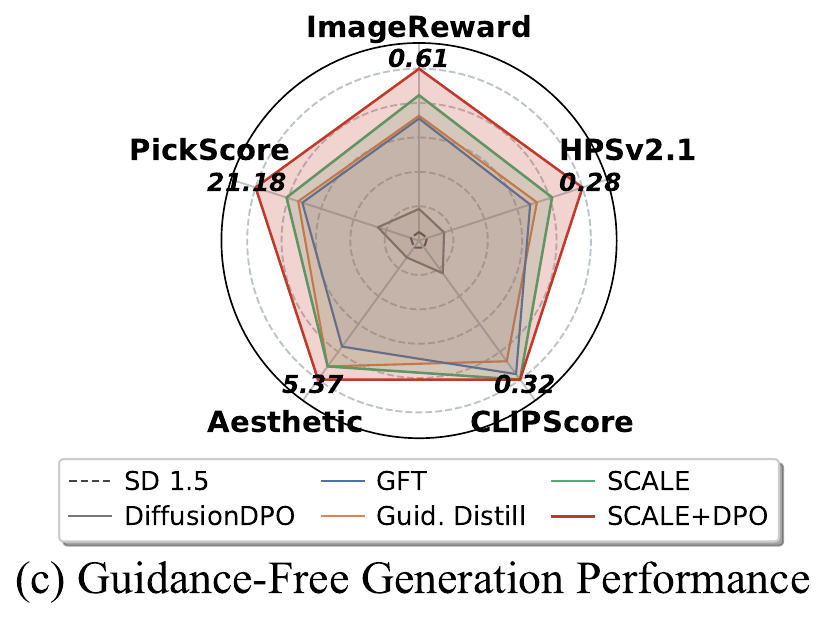}
    \vspace{-5mm}
  \end{subfigure}\hfill
  \caption{\textbf{Comparison between CFG and our self-contrastive guidance (SCG).} (a) CFG amplifies the conditional signal to improve sample quality, often reducing sample diversity by overemphasizing the condition. (b) SCG derives a model-based signal that measures the discrepancy between real and self-generated samples, pushing generation toward the true data manifold.
  (c) SCALE integrates CFG and SCG into a single model, achieving superior guidance-free generation performance. }
  \label{fig:concept}
\end{figure}

To eliminate this overhead, recent guidance-free approaches approximate the CFG distribution with a single inference via teacher distillation~\cite{distillation}, contrastive objectives~\cite{cca}, or direct reparametrization of the CFG objective~\citep{gft}.
While effective, these methods are fundamentally constrained to CFG and are bounded by its performance.
Furthermore, recent analyses have clarified that CFG is not flawless. A primary challenge is that condition alignment and sample fidelity are entangled along a single guidance. 
Therefore, amplifying the CFG for higher fidelity over-amplifies the conditional signal. This often leads to distorted sampling trajectories~\cite{chung2024cfg++}, greatly exaggerated truncation~\cite{autoguidance}, and mode dropping that collapses samples onto a narrow set of per-class templates~\cite{kynkaanniemi2024applying} (\cref{fig:concept}a). Whether a single-model approach can move beyond these limitations, therefore, remains an open question.



A complementary line of work, {information-based sampling guidance}~\cite{sag,hur2024unlocking} generalizes CFG by amplifying any salient information while avoiding a degraded reference that lacks it. While effective at targeting specific quality dimensions, these methods still incur additional sampling-time forward passes (if combined with CFG, the total cost steeply rises to three model evaluations per step) and rely on heuristic, domain-specific definitions of the salient information and the degraded reference.

In this paper, we propose \emph{self-contrastive guidance} (SCG), which contrasts the model's prediction on real and self-generated samples (\cref{fig:concept}b). 
This contrast yields a direct, model-based estimate of the model's own approximation error, and extrapolating along it pushes the generation process toward the true data manifold.
Unlike prior information-based guidance, SCG is domain-agnostic and is directly applicable to both diffusion and AR models as it relies only on model predictions.

By reparametrizing the SCG signal as a learnable term within the standard likelihood objective, we obtain \emph{Self-Contrastive Alignment for Likelihood Estimation} (SCALE), a guidance-free training framework in which the model learns to directly produce SCG-enhanced predictions in a single forward pass. For conditional generation, we further integrate CFG into SCALE so that the trained model directly refines the CFG distribution rather than simply mimicking it. 

Experimental results show that SCALE consistently surpasses existing guidance-free methods that only approximate CFG across diffusion and AR architectures, on both class-conditional ImageNet (DiT~\cite{dit}, LlamaGen~\cite{llamagen}) and text-to-image generation (Stable Diffusion v1.5~\cite{ldm}). It matches DPO-finetuned guidance-free models alone, and applying DPO on top of SCALE brings further gains (\cref{fig:concept}c). 
SCALE introduces minimal additional training overhead and complexity without compromising the base model's capabilities.
As a result, SCALE provides an orthogonal direction for improving guidance-free visual generation, widely applicable to both diffusion and AR models.

\section{Background} \label{sec:background}

\newcommand{\vepsilon}{\boldsymbol{\epsilon}}

\subsection{Likelihood-based Generative Models in Vision}
Likelihood-based generative models aim to explicitly model the data distribution $p(\vx)$. In this paper, we focus on two representative model types: diffusion~\cite{ho2020denoising} and autoregressive (AR)~\cite{llamagen} models.

\noindent
\textbf{Diffusion models}~\cite{ho2020denoising} aim to generate continuous data, where directly maximizing likelihood is intractable. These models learn to reverse a forward noising process that incrementally adds Gaussian noise $\vepsilon \sim \gN(\mathbf{0}, \mathbf{I})$ to a data $\vx$. The reverse process is learned via a score-matching objective~\cite{song2020score}, where a neural network $\vepsilon^c_{\theta}(\vx_t, t, c)$ with condition $c$ is trained to approximate the score function $\nabla_{\vx_t} \log p_t(\vx_t)$ through noise prediction from noisy sample $\vx_t = \alpha_t \vx + \sigma_t \vepsilon$ at timestep $t \in [0,T]$:
\begin{equation}
\label{eq:dm_loss}
\mathcal{L}_\theta^\text{DM} = \mathbb{E}_{(\vx_0,c) \sim p_{\text{data}}, \vepsilon, t } \left[ || \vepsilon - \vepsilon^c_{\theta}(\vx_t, t, c) ||^2 \right],
\end{equation}
where $\alpha_t$ and  $\sigma_t$ are scheduling functions satisfying $\alpha_t^2 + \sigma_t^2 = 1$, and $\vx_T \approx \vepsilon$, $\vx_0 \approx \vx$.
After training, the optimal noise predictor satisfies $\vepsilon^*_{\theta}(\vx_t) = -\sigma_t\nabla_{\vx_t} \log p_t(\vx_t)$, where $p_t$ denotes the marginal distribution at step $t$, enabling data generation via progressive denoising from $t=T$ to $0$.

\noindent
\textbf{Autoregressive models}~\citep{van2016pixel,vaswani2017attention} learn the joint distribution $p(\rvx,c) = \prod_{i=1}^{N} p^c(\vx_i | \vx_{<i}, c),$
where $\rvx = (\vx_1, \dots, \vx_N)$ is a discrete token sequence (e.g., vector-quantized image patches~\cite{vqvae2}) and $\vx_{<i}$ denotes all tokens preceding $\vx_i$. 
AR models are trained by minimizing the cross-entropy loss:
\begin{equation}
    \label{eq:ar_loss}
    \mathcal{L}_\theta^\text{AR} = \mathbb{E}_{(\rvx,c) \sim p_{\text{data}}} \left[ -\sum_{i=1}^{N} \log p^c_{\theta}(\vx_i | \vx_{<i}, c) \right].
\end{equation}

\paragraph{Classifier-free guidance.} \label{sec:guidance}

For complex visual datasets, finite-capacity networks often produce samples whose clarity falls short of the training images due to approximation errors. To mitigate this gap, likelihood-based visual generative models, including diffusion and AR models, commonly employ classifier-free guidance (CFG)~\cite{adm,ho2022classifier,chang2023muse}, which amplifies the conditional signal by extrapolating between the unconditional prediction $f^u_\theta$ and the conditional prediction $f^c_\theta$ with scale $s_\text{cfg}$:
\begin{equation}
    f_\theta^\text{cfg}(\vx_t,c) = f_\theta^c(\vx_t,c) + s_\text{cfg} \ (f_\theta^c(\vx_t,c)-f_\theta^u(\vx_t)).
    \label{eq:cfg}
\end{equation}
For a unified formulation, we let $\vx_t$ denote the model input at step $t$: the noised sample in diffusion models, or the preceding token sequence $\vx_{<t}$ in AR models.\footnote{We slightly abuse the notation $t$ to represent both inputs; $t$ denotes the timestep in diffusion, or the token index in AR.} Accordingly, $f_\theta$ is a parameterized network that serves as the noise predictor or the next-token predictor, respectively. Likewise, $p_t$ denotes the marginal at noise level $t$ in diffusion and the next-token distribution given $\vx_{<t}$ in AR.



\providecommand{\sg}{\mathbf{sg}} 
\subsection{Guidance-Free Visual Generation via Mimicking CFG}
Several approaches approximate the CFG distribution $p_t^{\text{cfg}}$ with a single inference, including guidance distillation~\citep{distillation} for diffusion models, Condition Contrastive Alignment (CCA)~\citep{cca} for AR models, and Guidance-Free Training (GFT)~\citep{gft} for both. GFT in particular integrates the CFG output directly into standard likelihood training by introducing a pseudo-temperature $\beta \coloneqq 1/(1+s_\text{cfg})$ and reformulating the conditional output $f_\theta^c$ in \eqref{eq:cfg} as an interpolation of CFG and unconditional predictions:
\begin{equation}
\begin{split}
    f^c_\theta(\vx_t,c,\beta) = \beta f^\text{cfg}_\theta(\vx_t,c,\beta) + (1-\beta) \text{sg}\left[f^u_\theta(\vx_t|\emptyset, 1)\right],
    \label{eq:gft_loss}
\end{split}
\end{equation}
where a stop-gradient $\sg[\cdot]$ is applied to keep the training cost minimal without affecting convergence.


Despite halving the inference cost of CFG, these methods remain constrained to the CFG-induced distribution $p_t^{\text{cfg}}$ as their training target, which alone is insufficient for high-quality generation due to its inherent limitations.
In the following section, we examine these limitations and motivate a complementary guidance signal that directly improves the guidance-free visual generation.

\section{Revisiting Information-Based Guidance for Guidance-Free Training} \label{sec:motivation}

\begin{figure*}[!t]
  \centering
    \includegraphics[width=\linewidth]{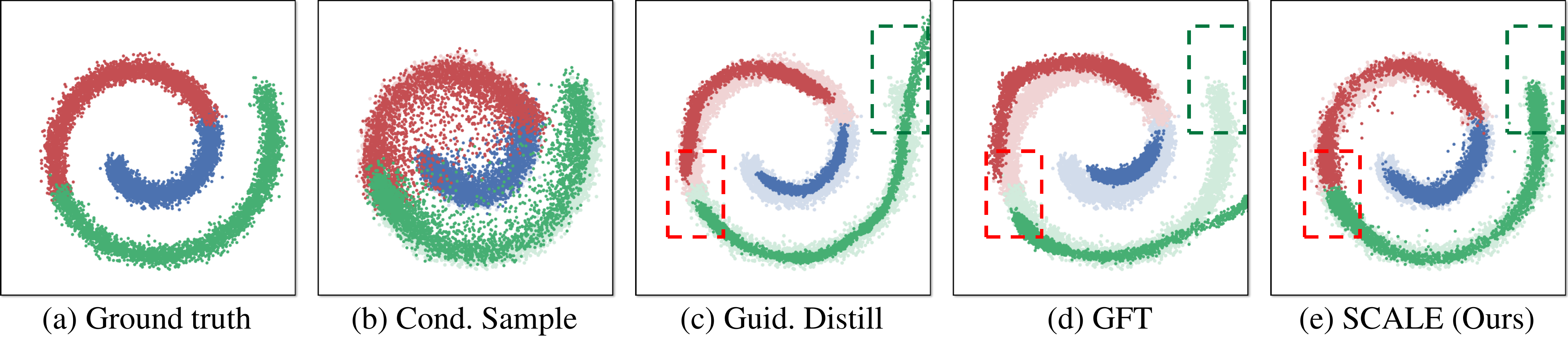}
    \caption{\textbf{Toy experiment on guidance-free generation with diffusion models.} 
    \textbf{(a)} Ground-truth samples from a three-class 2D Swiss-roll distribution. 
    \textbf{(b)} Unguided conditional sampling scatters outliers across low-density regions. 
    \textbf{(c, d)} Guidance-free methods that mimic CFG~\cite{distillation,gft} generate samples that drift away from class boundaries (red boxes) and off the data manifold (green boxes).
    \textbf{(e)} SCALE sharpens the distribution while preserving both the class boundary and the data manifold. The effective guidance scale is fixed at 5.0 across all guidance-free methods.}
  \label{fig:toy_exp}
\end{figure*}


\paragraph{CFG is not flawless.}
CFG enhances conditional generation by implicitly amplifying the conditional density to a power greater than one, sharpening the distribution toward its modes~\cite{bradley2024predictor}. While effective, recent analyses have clarified that CFG does not merely trade compute for quality: its unconditional-to-conditional extrapolation fundamentally entangles prompt alignment and perceptual quality, making them difficult to control independently~\cite{autoguidance}. 
Consequently, scaling up CFG for higher fidelity disproportionately amplifies the conditional signal, driving samples off the data manifold and truncating the distribution toward a narrow set of per-class ``templates''~\cite{kynkaanniemi2024applying,chung2024cfg++}.
Existing guidance-free generation methods halve the inference cost of CFG, but they do not address this underlying distortion.
%
We illustrate this in \cref{fig:toy_exp}. CFG-mimicking guidance-free methods over-amplify the conditioning signal, pulling samples away from class boundaries and off the data manifold (Fig.~\ref{fig:toy_exp}c and \ref{fig:toy_exp}d). 

\paragraph{Generalized Information-Based Guidance.}
To address these limitations, we first revisit a complementary line of work, information-based guidance~\cite{sag,hur2024unlocking}. It generalizes CFG by steering the sampling toward amplifying any salient information $h_t$ while avoiding degraded input samples $\Bar{\vx}_t$ lacking it. For diffusion models, this is formulated as
\begin{equation}
    \hat \vepsilon_\theta(\Bar{\vx}_t, h_t) = \vepsilon_\theta(\Bar{\vx}_t,h_t) + s \ (\vepsilon_\theta(\Bar{\vx}_t,h_t)-\vepsilon_\theta(\Bar{\vx}_t)),
\label{eq:sag_concise}
\end{equation}
and extends to discrete generative models from the optimization perspective\footnote{While \citet{hur2024unlocking} formulate this with $p_\theta(\tilde \vx_t|h_t)$, we instead use the joint $p_\theta(\tilde \vx_t, h_t)$. See Appendix~\ref{app:gen_info_guidance} for details.}:
\begin{equation}
    \argmax_{\bar\vx_t}[\log p_\theta(\Bar{\vx}_t,h_t) + s \ (\log p_\theta(\Bar{\vx}_t,h_t) - \log p_\theta(\Bar{\vx}_t))],
\label{eq:dsag_concise}
\end{equation}
where $s$ is a guidance scale. Both formulations play the same role of amplifying $h_t$ throughout the sampling process: \eqref{eq:sag_concise} minimizes the gradient of an implicit regressor $\nabla_{\Bar \vx_t} \log p_{\text{im}}(h_t|\Bar{\vx}_t)$ in the continuous score domain, whereas \eqref{eq:dsag_concise} directly amplifies the probability density $\log p_\theta(h_t|\Bar{\vx}_t)$. While these methods can target specific quality dimensions, they require additional sampling-time forward passes, and the total cost rises to three model evaluations per step if combined with CFG. Furthermore, they rely on heuristic and domain-specific choices for defining $\Bar{\vx}_t$ and $h_t$, such as applying a Gaussian blur with attention mask~\cite{sag} or utilizing models trained on auxiliary tasks~\cite{hur2024unlocking}. 

\paragraph{Our approach.}
Our key idea is to contrast the model's prediction on a real sample with its prediction on a self-generated sample. 
This contrast provides a direct model-based signal of the discrepancy between real and self-generated samples, and pushing generation along this direction encourages samples to move toward the data manifold. 
Because this signal depends on real samples, it is not available at sampling time, but it arises naturally during training. We therefore incorporate it into the guidance-free training objective so that the model learns to approximate the same guidance effect with a single forward pass at inference time (\cref{fig:toy_exp}e). Furthermore, since this formulation relies only on comparing model outputs, it applies broadly across likelihood-based generative models. In this paper, we specifically focus on two representative model types: diffusion and AR architecture.

\section{Self-Contrastive Alignment for Likelihood Estimation}
In this section, we formalize self-contrastive guidance (SCG) and integrate it into a guidance-free training objective, named Self-Contrastive Alignment for Likelihood Estimation (SCALE).
\subsection{Self-Contrastive Guidance} \label{sec:idg}
We formulate SCG by contrasting outputs from real and self-generated samples.
Formally, we define this by substituting a pair of real and generated samples $(\vx_t, \tilde{\vx}_t)$ into Eqs.~(\ref{eq:sag_concise}) and (\ref{eq:dsag_concise}), by setting $\Bar{\vx}_t = \tilde{\vx}_t$ and $h_t = \vx_t - \tilde{\vx}_t$ and assuming $f_\theta(\tilde \vx_t,\vx_t-\tilde \vx_t)=f_\theta(\vx_t)$:
\begin{equation}
    f^\text{scg}_\theta(\vx_t) = f_\theta(\vx_t) + s_\text{scg} \bigr(f_\theta(\vx_t) - f_\theta(\tilde \vx_t)\bigr),
    \label{eq:idg}
\end{equation}
where $s_\text{scg}$ controls the strength of the SCG.
This formulation applies to a broad class of generative models, including both continuous and discrete modalities.
By the definition of information-based guidance above, \eqref{eq:idg} steers the sampling trajectory toward enhancing $h_t = \vx_t - \tilde{\vx}_t$, locally amplifying the model's residual contrast between real and self-generated supports. This residual encodes how the model's prediction on a real sample diverges from its prediction on the self-generated sample, and extrapolating along it pushes the generation toward the high-density region of the data.

\begin{figure*}[!t]
    \centering
      \includegraphics[width=\linewidth]{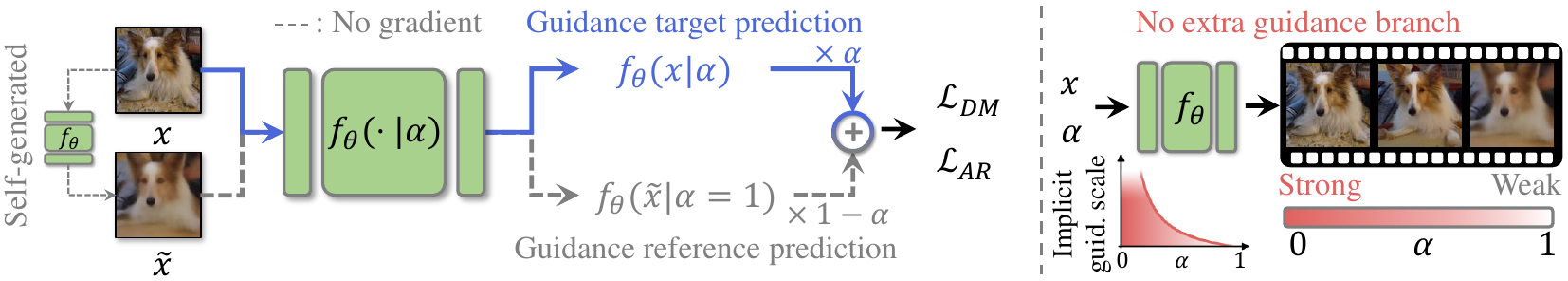}
    \caption{
        Illustration of SCALE algorithm without generation step $t$ for brevity. \textbf{Left:} SCALE training derives a model-based signal that measures the discrepancy between real ($\vx$) and self-generated ($\tilde\vx$) samples, learning to approximate the self-contrastive guidance with condition $\alpha$. 
        \textbf{Right:} At sampling time, lowering $\alpha$ yields high-quality samples by amplifying the implicit guidance signal.
        } 
        \label{fig:scale_pipeline}
\end{figure*}

\subsection{Guidance-Free Training with Self-Contrastive Guidance} \label{sec:train_idmix}

We integrate SCG into maximum likelihood training, so that the trained model directly approximates the self-contrastive signal and performs guidance-free sampling at sampling time.
Inspired by recent guidance-free training methods~\citep{gft}, we reparameterize the model output $f_\theta(\vx_t)$ as an interpolation between the target SCG output $f_\theta^\text{scg}(\vx_t)$ and the prediction from the self-generated sample $f_\theta(\tilde \vx_t)$:
\begin{equation}
    \label{eq:mug_reform}
    f_\theta(\vx_t) = \frac{1}{1+s_\text{scg}} 
    f_\theta^\text{scg}(\vx_t) + \frac{s_\text{scg}}{1+s_\text{scg}} f_\theta(\tilde \vx_t).
\end{equation}
To enable the model to generalize across various guidance strengths, we train the model $f_\theta(\vx_t | \alpha)$ conditioned on a pseudo-temperature $\alpha \coloneqq 1/(1+s_\text{scg})$ from \eqref{eq:mug_reform}:
\begin{equation}
    \label{eq:mug_gft}
    f_\theta(\vx_t \mid \alpha) = \alpha f_\theta^\text{scg}(\vx_t \mid \alpha) + (1-\alpha) \text{sg}[f_\theta(\tilde \vx_t \mid 1)].
\end{equation}
Here, $\alpha=1$ eliminates SCG ($s_\text{scg}=0$), while decreasing $\alpha$ amplifies the self-contrastive signal.
Substituting \eqref{eq:mug_gft} into standard likelihood objectives (Eqs.~(\ref{eq:dm_loss}) or (\ref{eq:ar_loss})) yields SCALE training objective.
To ensure the self-generated branch $f_\theta(\tilde \vx_{t} \mid 1)$ serves as a static reference for guidance, we apply a stop-gradient operator $\text{sg}[\cdot]$ and fix $\alpha=1$ on this branch.

We now characterize the optimal SCG prediction learned by SCALE training.


\begin{theorem}[Optimal Solution of SCALE Training] \label{prop:scale_optimality}
At an ideal regression optimum with unlimited model capacity and training data, assume the base predictor $\vepsilon_{\theta^*}(\cdot\mid1)$ matches the scaled data score at the real and reference inputs. For $\alpha\in(0,1]$, the SCG branch satisfies
\begin{equation}
\vepsilon^\text{scg}_{\theta^*}(\vx_t \mid \alpha) \;=\; -\sigma_t \left[ \frac{1}{\alpha} \nabla_{\vx_t} \log p_t(\vx_t) - \left(\frac{1}{\alpha} - 1\right) \mathbb E\bigl[\nabla_{\tilde \vx_t} \log p_t(\tilde \vx_t)\mid\vx_t\bigr] \right],
\label{eq:scale_optimal_form}
\end{equation}
where $\tilde{\vx}_t=g_{\theta^*}(\vx_t,\vepsilon)$ is obtained by estimating the clean sample with the base predictor and re-noising it to level $t$ using $\vepsilon$. The expectation is taken over $\vepsilon$.
\end{theorem}

We defer the proof and an analogous result for AR models to Appendix~\ref{app:proof_optimality}. Since $\alpha = 1/(1+s_\text{scg})$, the SCALE-trained diffusion model outputs the conditional mean SCG prediction in a single forward pass at convergence, with guidance strength controlled by $\alpha$. \cref{fig:scale_pipeline} illustrates the SCALE training and sampling without external condition $c$.

\subsection{Practical Implementation} \label{sec:mixup}


\paragraph{Regularizing SCALE via interpolation.} \label{sec:practical_implementation}

The analysis above assumes that $\vx_t$ and $\tilde \vx_t$ lie on nearby supports so that the prediction $f_\theta(\tilde \vx_t)$ is well defined and the local density-ratio is informative. In practice, a self-generated sample can be highly degraded under limited model capacity or the stochasticity of early generation steps ($t \ll T$), making direct contrast unstable~\cite{autoguidance,hyung2025spatiotemporal}. 

To address this, we regularize SCALE training by interpolating the real and self-generated samples into an intermediate support, thereby providing a stable anchor for the guidance.
Concretely, we replace a fraction of patches in $\vx_t$ with those from $\tilde \vx_t$. Given an interpolation ratio $\lambda \in [0,1]$ and a per-patch Bernoulli mask $M_\lambda \sim \mathrm{Bernoulli}(\lambda)$, the mixed sample $\tilde \vx_{t,\lambda}$ is defined as
\begin{equation}
\tilde \vx_{t,\lambda} :=  \text{PatchMix}(\vx_t, \tilde \vx_t; \lambda) = M_\lambda \odot \tilde \vx_t + (1-M_\lambda) \odot \vx_t,
\label{eq:mixup}
\end{equation}
where $\odot$ denotes patch-wise multiplication. This preserves local semantic structures of dense visual data and applies uniformly across both continuous and discrete data representations, unlike linear interpolation, which is ill-defined for vector-quantized tokens (see Appendix~\ref{app:mixup_visualize}). 
With this interpolated output, the SCG and SCALE formulations in \eqref{eq:idg} and \eqref{eq:mug_gft} can be redefined as
\begin{gather}
f^\text{scg}_\theta(\vx_t) = f_\theta(\vx_t) + s_\text{scg}(f_\theta(\vx_t) - f_\theta(\tilde \vx_{t,\lambda})).
\label{eq:mug}\\
f_\theta(\vx_t \mid \alpha) = \alpha f_\theta^\text{scg}(\vx_t \mid \alpha) + (1-\alpha) \text{sg}[f_\theta(\tilde \vx_{t,\lambda} \mid 1)].
\label{eq:mug_gft_patchmix}
\end{gather}

\paragraph{Reference choice and training stability.}
In \eqref{eq:mug_gft_patchmix}, we utilize the unguided reference from the $\alpha=1$ slice with stop-gradient and PatchMix regularization. 
This stabilizes the SCALE training by keeping target dependencies acyclic and reducing the effects of reference displacement. We analyze these in the diffusion setting.
\textit{(i) Unguided reference.} Stop-gradient applies only to the reference branch; the same network is trained through the prediction branch with $\alpha$ sampled uniformly from $[\alpha_{\min},1]$. Assuming the predictor is smooth in $\alpha$ near $1$, training at nearby values also constrains its $\alpha=1$ slice, whose regression target is fixed by the data. The resulting chain of target definitions is acyclic within each update (data $\to\alpha=1\to\alpha<1$). As the $\alpha=1$ slice converges, the SCG target stabilizes, and the $\alpha<1$ slices can converge toward it.
\noindent\textit{(ii) PatchMix.} Interpolation brings the reference closer to the real input on average. For finite-capacity models, PatchMix reduces the contributions from reference displacement to the bounds on SCG target variation and prediction error by a factor of $\sqrt{\lambda}$. These bounds support using mixed samples as more reliable guidance anchors when self-generated samples are degraded. Detailed proofs are provided in Appendix~\ref{app:scale_training_analysis}.


\begin{figure*}[t]
    \begin{minipage}[t]{0.47\textwidth}
    \vspace{0pt}  
    \hrule height 0.8pt \vspace{3pt}
    \captionof{algorithm}{SCALE Training (Diffusion)}
    \label{alg:mug_diff}
    \hrule \vspace{3pt}
    \begin{algorithmic}[1]
        \STATE Initialize denoiser $\vepsilon_\theta$ and ratio $\lambda$ 
        \FOR{each gradient step} 
        \STATE $\vx, c \sim p(\vx, c)$, $\vepsilon \sim \gN(\mathbf{0}, \mathbf{I})$,\\
               $t, \alpha \sim U(\alpha_{\min},1)$, $\beta \sim U(0,1)$
        \STATE $\vx_t = \alpha_t \vx + \sigma_t \vepsilon$
        \STATE $\textcolor{blue}{\vepsilon_\theta^\text{{scg-cfg}} = \vepsilon_\theta(\vx_t|c,\alpha,\beta)}$ \hfill \small{\color{blue} \Comment{Target}}
        \STATE \small{\color{gray} \textit{// Additional to Training (Evaluation Mode)}} 
        \STATE $\vepsilon_\theta^u = \vepsilon_\theta(\vx_t|\varnothing,1,1)$ 
        \STATE Sample $\tilde \vx_t$ from $\vepsilon_\theta^u$ 
        \STATE $\tilde \vx_{t,\lambda} = \text{PatchMix}(\vx_t, \tilde \vx_t; \lambda)$ \hfill \Comment{\cref{eq:mixup}} 
        \STATE $\vepsilon_\theta^\lambda = \vepsilon_\theta(\tilde \vx_{t,\lambda}|c,1,\beta)$ 
        \STATE \small{\color{gray} \textit{// Nested guidance formulation}} \hfill \Comment{\cref{eq:finalv2}} 
        \STATE $\textcolor{blue}{\vepsilon_\theta^\text{cfg} = \alpha \, \vepsilon_\theta^\text{{scg-cfg}} + (1{-}\alpha) \,\text{sg}[\vepsilon_\theta^\lambda]}$
        \STATE $\textcolor{blue}{\vepsilon_\theta^c = \beta \, \vepsilon_\theta^\text{cfg} + (1{-}\beta) \,\text{sg}[\vepsilon_\theta^u]}$ 
        \STATE $\theta \leftarrow \theta - \mu \nabla_\theta \| \vepsilon - \textcolor{blue}{\vepsilon_\theta^c} \|^2_2$ \hfill \Comment{\cref{eq:dm_loss}}
        \ENDFOR
    \end{algorithmic}
    \vspace{3pt} \hrule height 0.8pt
    \end{minipage}
    \hfill
    \begin{minipage}[t]{0.5\textwidth}
        \vspace{-1pt}
        \captionof{table}{\textbf{Comparison of guidance-free generation methods.} Computational costs are reported relative to CFG for both training and sampling.
        {SCALE} keeps the low training overhead while halving inference cost and improving the target distribution alignment. 
        }
        \vspace{2pt}
        \centering
        \small
        \setlength{\tabcolsep}{2.8pt}
        \begin{tabular}{l|cccc}
        \toprule
        Method     & Distill   & CCA    & GFT           & \textbf{SCALE} \\
        \midrule
        Guidance target & $\frac{p_t(\vx|c)}{p_t(\vx)}$ & $\frac{p_t(\vx|c)}{p_t(\vx)}$ & $\frac{p_t(\vx|c)}{p_t(\vx)}$ & $\frac{p_t^\text{cfg}(\vx|c)}{p_t^\text{cfg}(\tilde\vx|c)}$ \vspace{1mm} \\
        Applicable to & \scriptsize Diffusion & \makecell{\scriptsize Discrete} & \scriptsize All  & \scriptsize All \vspace{1mm} \\
        \midrule
        \makecell[l]{Train time \\ / Step} & $\times 1.12$ & $\times 1.69$ & $\times 1.06$ & $\times 1.12$ \vspace{1mm} \\
        Params     & $\times 1.00$ & $\times 1.00$ & $\times 1.00$ & $\times 1.00$ \vspace{1mm} \\
        VRAM       & $\times 1.21$ & $\times 1.39$ & ${\times 1.00}$ & ${\times 1.00}$ \vspace{1mm} \\
        Inference  & $\times 0.50$ & $\times 0.50$ & ${\times 0.50}$ & ${\times 0.50}$ \vspace{1mm} \\
        \bottomrule
        \end{tabular}
        \label{tab:comparison}
    \end{minipage}
\end{figure*}

\paragraph{SCALE for conditional generation.}
The basic SCALE objective sharpens sample distribution but does not enforce condition alignment, which becomes critical in large-scale visual generation. We therefore integrate CFG into SCALE training, while ensuring the inference fully guidance-free.
Formally, we apply regularized SCG (\cref{eq:mug}) on top of the CFG prediction $f^\text{cfg}_\theta$, which means it directly refines CFG-generated outputs using the self-contrastive signal:
\begin{equation}
\begin{split}
    f^\text{{scg-cfg}}_\theta(\vx_t \mid c) = f_\theta^\text{cfg}(\vx_t \mid c) + s_\text{scg}(f^\text{cfg}_\theta(\vx_t \mid c) - f^\text{cfg}_\theta(\tilde \vx_{t,\lambda} \mid c)).
\end{split}
\label{eq:mug_cfg}
\end{equation}
Following the derivation in \cref{sec:train_idmix}, we reformulate this unified target into a learnable form:
\begin{equation}
    \begin{split}
        f_\theta^\text{cfg}(\vx_t|c,\alpha) = \textcolor{blue}{\alpha f_\theta^{\text{{scg-cfg}}}(\vx_t|c, \alpha)
        + (1-\alpha) \ \text{sg}\left[f_\theta^\text{cfg}(\tilde \vx_{t,\lambda} \mid c, 1)\right]}.
    \end{split}
    \label{eq:final}
\end{equation}
Since $f_\theta^\text{cfg}$ on the left-hand side follows $p_t^\text{cfg}$, it is not directly learnable. We therefore integrate it into the guidance-free training framework for CFG (\eqref{eq:gft_loss}), yielding a fully tractable training target:
\begin{equation}
\begin{split}    
    &f^c_\theta(\vx_t|c, \alpha, \beta) = \\
    &\beta \left\{\textcolor{blue}{\alpha f_\theta^{\text{{scg-cfg}}}(\vx_t \mid c, \alpha,\beta) + (1-\alpha) \text{sg}\left[f_\theta^\text{cfg}(\tilde \vx_{t,\lambda} \mid c, 1, \beta)\right]} \right\} + (1-\beta) \ \text{sg}[f^u_\theta(\vx_t|\emptyset, 1, 1)],
\end{split}
\label{eq:finalv2}
\end{equation}
where $\beta$ is a hyperparameter controlling the CFG scale ($\beta=1/(1+s_\text{cfg})$).
This formulation seamlessly integrates SCG with CFG, allowing the target trained model $f_\theta^{\text{scg-cfg}}$ to be directly optimized by replacing the model prediction in Eqs.~\ref{eq:dm_loss} and~\ref{eq:ar_loss} with $f^c_\theta(\vx_t|c, \alpha, \beta)$.
Setting $\beta=1$ recovers basic SCALE, while setting $\alpha=1$ reduces to the GFT~\cite{gft}.
Consequently, the trained model generates samples that incorporate both SCG and CFG in a single forward pass.
In practice, we sample $\tilde \vx_t$ from the model's unconditional prediction during training. This adds no extra cost and yields a stable reference, free from both CFG and SCG amplification.
\cref{alg:mug_diff} outlines the SCALE training pipeline for diffusion models. The SCALE training pipeline for AR models is in Appendix~\ref{app:ar_scale}.

\subsection{Comparison with Other Methods} \label{sec:connection}
We compare SCALE against related methods; see Appendix~\ref{app:related} for more related works.

\paragraph{Computational overhead.}
As summarized in \cref{tab:comparison}, the training-time overhead is minimal and VRAM is unchanged relative to CFG training as the extra forward passes introduced by SCALE are computed without gradients. 
Since SCALE introduces minimal additional parameters for embedding $\alpha$ and $\beta$, the parameter ratio is unchanged.
We defer the detailed setup and analysis to Appendix~\ref{app:computational}.

\paragraph{Versus CFG-mimicking methods.}
Prior guidance-free methods train the model to reproduce the CFG distribution $p_t^{\text{cfg}}$, so their quality is bounded by CFG itself. SCALE instead amplifies a model-based density ratio, $p_t(\vx_t)/p_t(\tilde \vx_t)$ for basic SCALE or $p^\text{cfg}_t(\vx_t|c)/p^\text{cfg}_t(\tilde \vx_t|c)$ for CFG-augmented SCALE, providing a stronger alignment with the data distribution and superior sample quality.

\paragraph{Versus RL post-training.}
RL post-training methods such as DPO~\cite{rafailov2023direct} and GRPO~\cite{shao2024deepseekmath} also use a contrastive signal but pursue a different goal. They align the model to \emph{human preferences} via preference data or reward models and typically still rely on CFG in the visual domain~\cite{wallace2024diffusion,liu2025flow}, whereas SCALE aligns to the \emph{ground-truth data distribution} from the model's own predictions and is fully guidance-free. The two are therefore complementary, and \cref{par:scale_vs_dpo} shows that DPO~\cite{wallace2024diffusion} on top of SCALE further improves quality while remaining guidance-free.
\section{Results}

\subsection{Experimental Setups}
\paragraph{Models, datasets, and metrics.}
We evaluate SCALE on class-conditional (C2I) ImageNet~\cite{deng2009imagenet} at $256\times256$ by fine-tuning DiT-XL/2~\cite{dit} and LlamaGen-L~\cite{llamagen}, and additionally train DiT-B/2 for the analysis. We report FID-50k, IS, Precision, and Recall following \citet{adm}.
For text-to-image generation (T2I) at $512\times512$, we fine-tune Stable Diffusion v1.5 (SD1.5)~\cite{ldm} on LAION-Aesthetics 5+~\cite{schuhmann2022laion} and report FID-30k, CLIP Score~\cite{radford2021learning}, PickScore~\cite{pickscore}, HPSv2.1~\cite{hpsv2}, Aesthetic Score~\cite{schuhmann2022laion}, and Image Reward~\cite{imagereward}, complemented by a human preference study. Computational cost is measured by the number of function evaluations (NFE). See Appendix~\ref{sec:detail} for more details.

\paragraph{Baseline methods.}
We compare {SCALE} against three groups of baselines, which enhance the base model's performance in different ways.
\textit{Guidance-free methods}: We include guidance distillation~\cite{distillation} for diffusion models, CCA~\cite{cca} for AR models, and GFT~\cite{gft} for both.
\textit{Guided sampling methods}: We compare against CFG~\cite{ho2022classifier}, SAG~\cite{sag}, and discriminator guidance (G++)~\cite{discg}.
For text-to-image, we further compare against RL post-training with DiffusionDPO~\cite{wallace2024diffusion}.
We additionally report with representative diffusion models~\cite{adm,ldm,cdm}, and AR models \cite{vqgan,vit-vqgan,rq,var}.

\paragraph{Implementation details.}
We train all models with the SCALE objective in~\eqref{eq:finalv2}, introducing only a few additional hyperparameters. For fine-tuning, we limit training to less than $5\%$ of the pre-training budget and inherit the base model's hyperparameters. We fix the PatchMix ratio at $\lambda=0.2$ across all experiments. The patch size is matched to each model's self-attention granularity: $2$ for DiT, $1$ for LlamaGen, and $2$ for SD1.5 (the finest self-attention resolution in its UNet). For diffusion sampling, we use 50-step DPM-Solver++~\cite{lu2022dpmpp}.
See Appendix~\ref{sec:detail} for the full setup.

\begin{table}[t]
\caption{Comparisons of guidance-free class-conditional generation on ImageNet $256\times256$. \textbf{Bold} for the best and \underline{underline} for the second-best metrics for each model type. }
\label{tab:finetune_fid}
\centering
\small
\resizebox{.9\textwidth}{!}{
\begin{tabular}{c|l|cccc}
\toprule
\textbf{Model Types} & \multicolumn{1}{c|}{\textbf{Methods}} & \textbf{FID-50k} $\downarrow$ & \textbf{IS} $\uparrow$ & \textbf{Precision} $\uparrow$ & \textbf{Recall} $\uparrow$ \\
\midrule
\multirow{7}{*}{\shortstack[c]{Diffusion\\Models}}
& ADM~~\citep{adm}& 10.94  & 101.1 & 0.69 & \underline{0.63} \\
& LDM-4~~\citep{ldm}& 10.56 & 103.4 & \underline{0.72} & 0.62 \\
& CDM~\cite{cdm} & 4.88  & 158.7  & --  & -- \\
& DiT-XL/2~~\citep{dit} & 9.62 & 121.5 & 0.67 & \textbf{0.67} \\
& $\quad+$Guid. Distill~~\citep{distillation}    & 2.11   & -- & -- & -- \\
& $\quad+$GFT ~\citep{gft}    & \underline{1.99}  & \underline{266.6} & \textbf{0.81} & 0.60 \\
& \cellcolor[RGB]{238, 238, 228}$\quad+$\textbf{{SCALE} (Ours)}    & \cellcolor[RGB]{238, 238, 228}\textbf{1.87} & \cellcolor[RGB]{238, 238, 228}\textbf{268.0} & \cellcolor[RGB]{238, 238, 228}\textbf{0.81} & \cellcolor[RGB]{238, 238, 228}0.60 \\
\midrule
\multirow{8}{*}{\shortstack[c]{Autoregressive\\Models}}
& VQGAN~~\citep{vqgan}& 18.65 & 80.4 & 0.78 & 0.26 \\
& ViT-VQGAN~~\citep{vit-vqgan}& 4.17 & 175.1 & -- & -- \\
& RQ-Transformer~~\citep{rq} & 7.55  & 134.0 & -- & -- \\
& VAR-d30~~\citep{var}        & 5.26 & -- & -- & -- \\
& LlamaGen-L~~\citep{llamagen} & 19.0  & 61.7 & 0.60 & \textbf{0.66} \\
& $\quad+$CCA~~\citep{cca}    & 3.43  & \textbf{288.2} & 0.81 & 0.52 \\
& $\quad+${GFT~\citep{gft}}   & \underline{2.88} & 238.4 & --  & -- \\
& \cellcolor[RGB]{238, 238, 228}$\quad+$\textbf{{SCALE} (Ours)}   & \cellcolor[RGB]{238, 238, 228}\textbf{2.58} & \cellcolor[RGB]{238, 238, 228}\underline{253.0} & \cellcolor[RGB]{238, 238, 228}\textbf{0.82} & \cellcolor[RGB]{238, 238, 228}\underline{0.55} \\
\bottomrule
\end{tabular}
}
\end{table}


\subsection{Enhancing Guidance-free Generation with SCALE}
In this section, we demonstrate that incorporating self-contrastive signal effectively enhances guidance-free generation by comparing SCALE with various generative models without guidance.

\begin{table}[t]
  \caption{Guidance-free generation results on text-to-image generation using SD1.5.}
  \label{tab:sd15_t2i}
  \centering
  \small
  \setlength{\tabcolsep}{3pt}
  \resizebox{.9\textwidth}{!}{
  \begin{tabular}{lcccccc}
  \toprule
  \textbf{Methods} & \textbf{PickScore} $\uparrow$ & \textbf{CLIP} $\uparrow$ & \textbf{HPSv2.1} $\uparrow$ & \textbf{FID-30k} $\downarrow$ & \textbf{Aesthetic} $\uparrow$ & \textbf{Img. Reward} $\uparrow$ \\
  \midrule
  SD 1.5~\cite{ldm} & 18.90 & 0.252 & 0.187 & 22.47 & 4.97 & -1.30 \\
  $\quad+$Guid. Distill~\cite{distillation} & 20.56 & 0.314 & 0.252 & 15.12 & \textbf{5.33} & 0.06 \\
  $\quad+$GFT~\cite{gft} & 20.50 & 0.321 & 0.248 & \textbf{12.50} & 5.27 & 0.02 \\
  \rowcolor[RGB]{238, 238, 228}
  $\quad+$\textbf{{SCALE} ($\alpha=0.8$)} & \underline{20.58} & \textbf{0.324} & \underline{0.253} & \underline{12.83} & \underline{5.30} & \underline{0.18} \\
  \rowcolor[RGB]{238, 238, 228}
  $\quad+$\textbf{{SCALE} ($\alpha=0.25$)} & \textbf{20.73} & \textbf{0.324} & \textbf{0.261} & 16.52 & \textbf{5.33} & \textbf{0.30} \\
  \bottomrule
  \end{tabular}
  }
  \vspace{-2mm}
\end{table}

  \begin{figure}[!ht]
    \centering
    \begin{subfigure}[b]{0.238\linewidth}
      \raisebox{2mm}{\includegraphics[width=\linewidth]{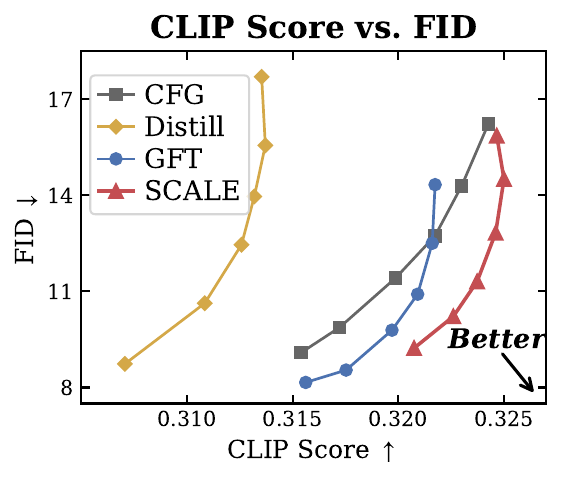}}
      \caption{Pareto curves.} \label{fig:sd15_t2i_pareto}
    \end{subfigure}\hfill
    \begin{subfigure}[b]{0.202\linewidth}
      \includegraphics[width=\linewidth]{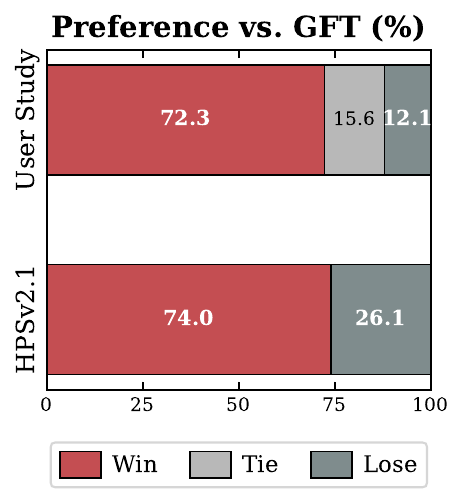}
      \caption{Win rate vs. GFT.} \label{fig:sd15_t2i_winrate}
    \end{subfigure}
    \begin{subfigure}[b]{0.55\linewidth}
      \includegraphics[width=\linewidth]{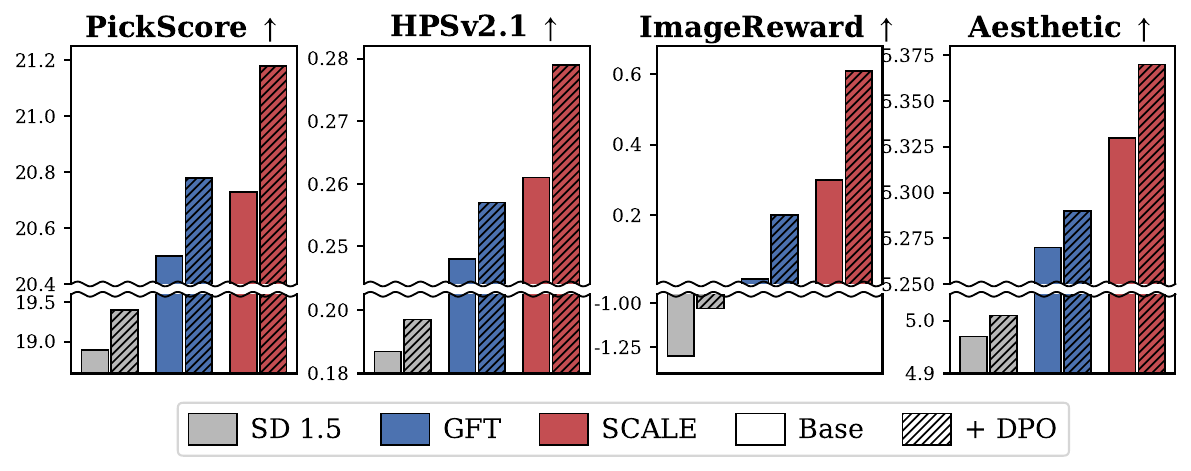}
      \caption{Compatibility with DPO~\cite{wallace2024diffusion}.} 
      \label{fig:sd15_t2i_dpo}
    \end{subfigure}\hfill
    \caption{Performance comparison of {SCALE} and other guidance-free baselines using SD 1.5~\cite{ldm}.}
  \end{figure}

  \begin{figure}[!t]
    \centering
    \includegraphics[width=\linewidth]{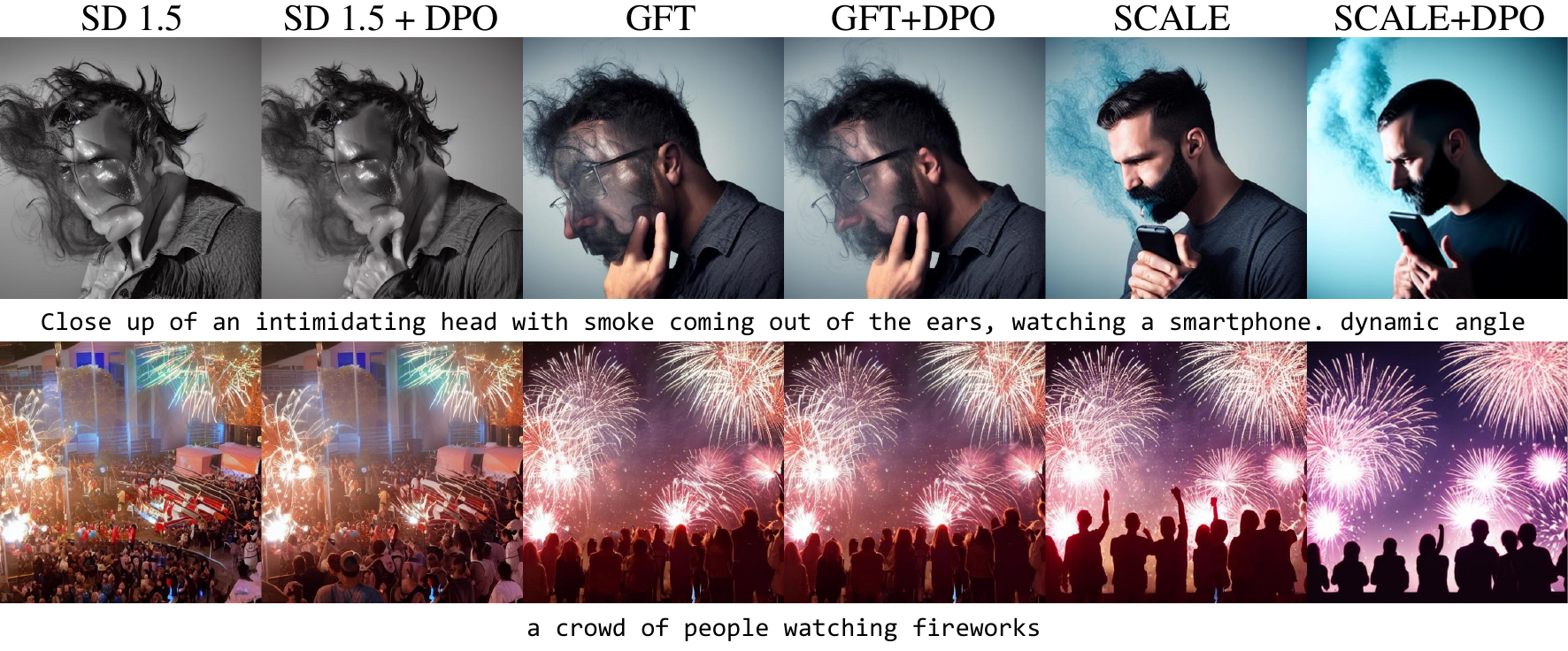}
    \caption{Guidance-free generation results of base SD 1.5~\cite{ldm}, GFT, and SCALE, with and without DPO post-training~\cite{wallace2024diffusion}. Prompts are sampled from the Pick-a-Pic test set~\cite{pickscore} and PartiPrompts~\cite{partiprompts}.}
    \label{fig:visualization_dpo_main}
    \vspace{-3mm}
\end{figure}

\paragraph{Fine-tuning from pretrained models (C2I).}
We compare SCALE against baseline guidance-free methods with various generative models on class-conditional generation on ImageNet $256\times256$. Among these, guidance distillation~\cite{distillation} applies only to diffusion models, and CCA~\cite{cca} only to AR models. In contrast, SCALE and GFT apply to both. As shown in \cref{tab:finetune_fid}, SCALE outperforms all baselines on FID and IS across both diffusion (DiT-XL/2~\cite{dit}) and AR (LlamaGen-L~\cite{llamagen}) models.

\paragraph{Fine-tuning from pretrained models (T2I).}
\cref{tab:sd15_t2i} compares guidance-free generation results on SD1.5. 
The effective CFG scale for guidance-free methods is set to the standard CFG scale of 7.5.
SCALE achieves superior performance while maintaining the competitive FID scores, and further improves the quality by lowering the $\alpha$, confirming that the self-contrastive signal generalizes to large-scale zero-shot generation. 
Lowering $\alpha$ compensates FID but further improves the quality. The Pareto curves in \cref{fig:sd15_t2i_pareto}, obtained by sweeping the effective CFG scale from 4.0 to 12.5 at $\alpha=0.8$, show that SCALE achieves a superior CLIPScore-FID trade-off. The head-to-head comparison in \cref{fig:sd15_t2i_winrate} further shows SCALE winning over GFT on both HPSv2.1 and PickScore by clear margins.

\paragraph{SCALE vs. DPO.} \label{par:scale_vs_dpo}
As discussed in \cref{sec:connection}, SCALE and DPO operate along different perspectives. 
Experimental results in \cref{fig:sd15_t2i_dpo} demonstrate three key findings:
\textit{(1)} In the guidance-free setting, DPO barely improves over the SD\,1.5 baseline, showing that DPO itself relies heavily on CFG for high-quality generation.
\textit{(2)} SCALE alone matches GFT+DPO, suggesting that self-contrastive signals are an effective alignment objective without preference data.
\textit{(3)} SCALE+DPO yields further gains across all metrics, indicating that the two signals are complementary rather than redundant.

The qualitative comparisons in \cref{fig:visualization_dpo_main} further support these findings. Without sampling-time guidance, SCALE produces samples with finer details and higher visual quality than base SD 1.5 and GFT. Applying DPO to SCALE further enhances these qualities, confirming the complementary benefits of self-contrastive signal and preference-based post-training. More qualitative comparisons of guidance-free generation with DPO post-trained models are in \cref{fig:visualization_dpo,fig:visualization_dpo_alpha_sweep} of the Appendix.




 \begin{table}[!t]
    \caption{Comparison with guided sampling methods on the ImageNet $256\times256$. $^\dagger$ Re-evaluated using 50-step DPM-Solver++~\cite{lu2022dpmpp} for a fair comparison. See Appendix~\ref{app:guided_ddpm} for more details.}
    \label{tab:guided}
    \centering
    \scriptsize
    \setlength{\tabcolsep}{4pt}
    \renewcommand{\arraystretch}{0.85}
    \resizebox{.9\textwidth}{!}{
    \begin{tabular}{clcccccc}
      \toprule
      \textbf{Models} & \textbf{Methods} & \textbf{Guidance} & \textbf{NFEs} $\downarrow$ & \textbf{FID} $\downarrow$ & \textbf{IS} $\uparrow$ & \textbf{Precision} $\uparrow$ & \textbf{Recall} $\uparrow$ \\
      \midrule
      \multirow{4}{*}{DiT-XL/2~\citep{dit}} & CFG$^\dagger$~\citep{ho2022classifier} & Required & 50 $\times$ 2 & 2.17 & \underline{243.7} & \underline{0.80} & 0.60 \\
      & SAG$^\dagger$~\citep{sag} & Required & 50 $\times$ 3 & 2.17 & 241.2 & \underline{0.80} & 0.59 \\
      & G++$^\dagger$~\citep{discg} & Required & 50 $\times$ 3 & \underline{2.14} & {240.0} & 0.76 & \underline{0.63} \\
      & \cellcolor[RGB]{238, 238, 228}\textbf{{SCALE} (Ours)} & \cellcolor[RGB]{238, 238, 228}{Free} & \cellcolor[RGB]{238, 238, 228}50 & \cellcolor[RGB]{238, 238, 228}\textbf{1.87} & \cellcolor[RGB]{238, 238, 228}\textbf{268.0} & \cellcolor[RGB]{238, 238, 228}\textbf{0.81} & \cellcolor[RGB]{238, 238, 228}0.60 \\
      \midrule
      \multirow{2}{*}{LlamaGen-L~\citep{llamagen}} & CFG~\citep{ho2022classifier} & Required & 256 $\times$ 2 & \underline{3.07} & \textbf{256.0} & \textbf{0.83} & 0.52 \\
      & \cellcolor[RGB]{238, 238, 228}\textbf{{SCALE} (Ours)} & \cellcolor[RGB]{238, 238, 228}{Free} & \cellcolor[RGB]{238, 238, 228}256 & \cellcolor[RGB]{238, 238, 228}\textbf{2.58} & \cellcolor[RGB]{238, 238, 228}\underline{253.0} & \cellcolor[RGB]{238, 238, 228}\underline{0.82} & \cellcolor[RGB]{238, 238, 228}\underline{0.55} \\
      \bottomrule
    \end{tabular}
    }
  \end{table}
  
\begin{figure}[!t]
  \centering
  \begin{subfigure}[b]{0.465\linewidth}
    \centering
    \includegraphics[width=\linewidth]{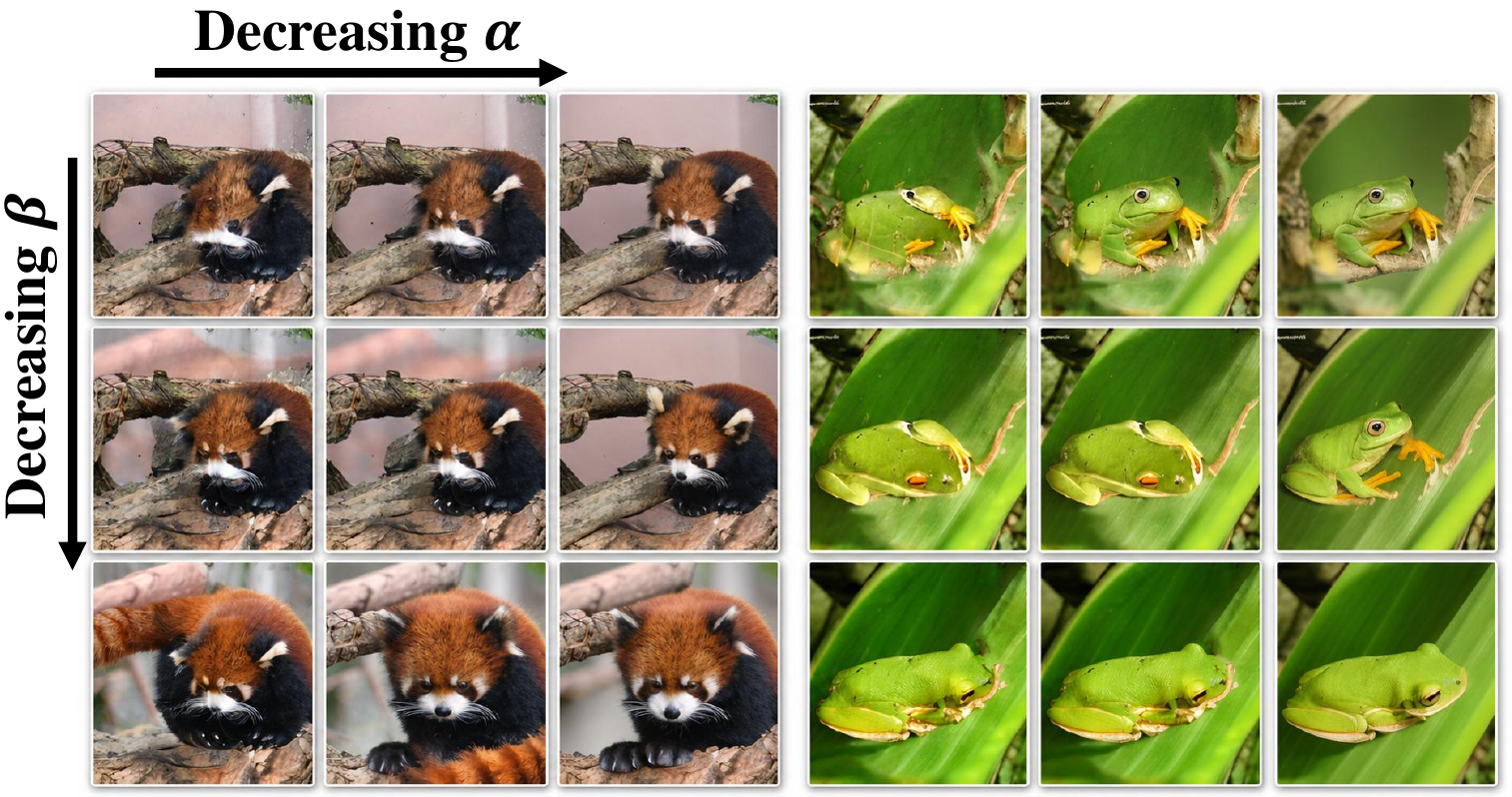}
    \caption{Generated samples.}
    \label{fig:ab_qualitative}
  \end{subfigure}\hfill
  \begin{subfigure}[b]{0.535\linewidth}
    \centering
    \includegraphics[width=\linewidth]{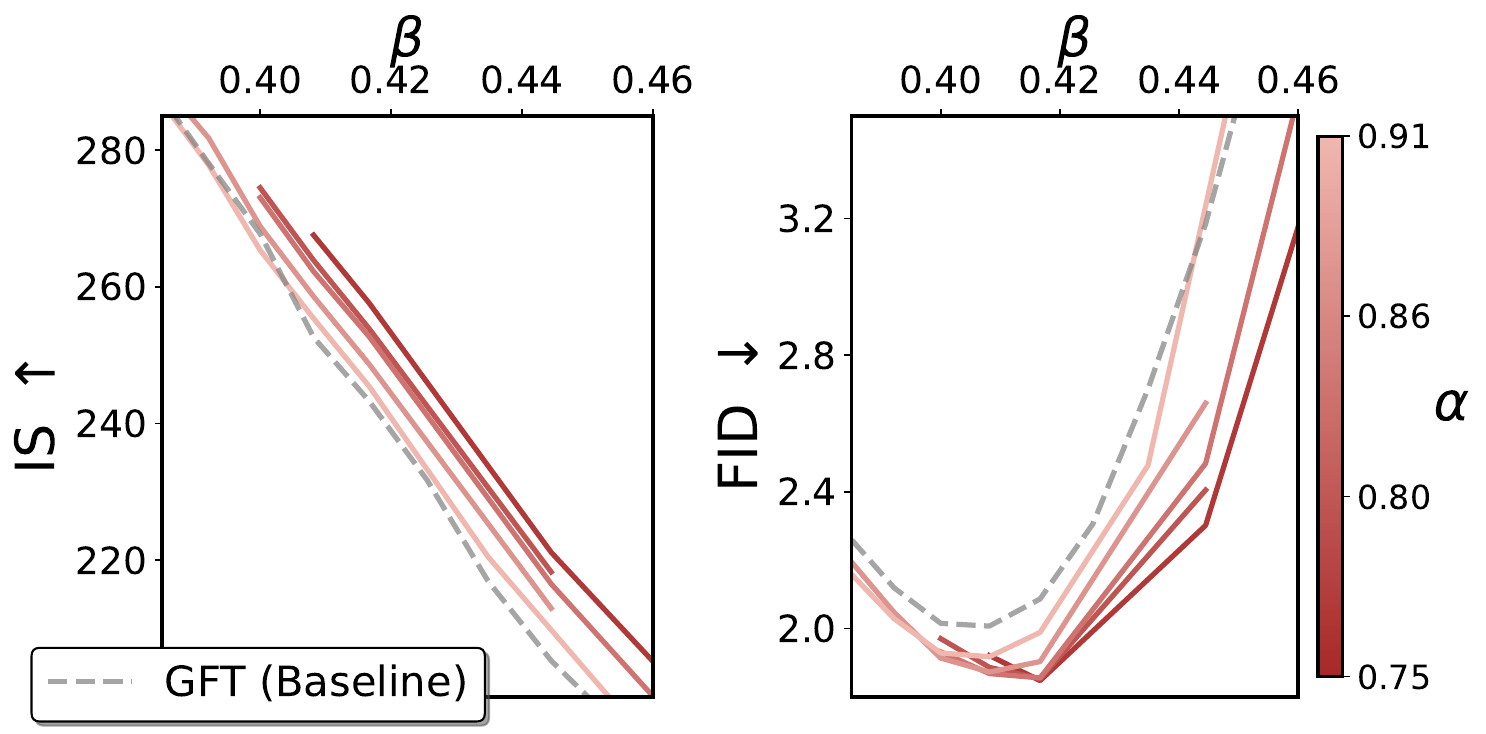}
    \caption{FID and IS curves.}
    \label{fig:ab_quantitative}
    \vspace{-1mm}
  \end{subfigure}
  \caption{Effect of varying $\alpha$ and $\beta$ of {SCALE} using DiT-XL/2~\cite{dit} at sampling time.}
  \label{fig:ab_analysis}
\end{figure}

\paragraph{Comparison with guided sampling.}
We compare SCALE against CFG~\cite{ho2022classifier}, information-based guidance for diffusion models (SAG~\cite{sag}), and guidance that utilizes an external discriminator (G++)~\cite{discg}.
As shown in \cref{tab:guided}, while these methods introduce additional sampling overhead, SCALE matches them without any guidance at sampling, and even surpasses CFG in FID.


\subsection{Analysis on {SCALE}} \label{sec:analysis}

\paragraph{Effect of varying $\alpha$ and $\beta$.} 
As noted in \cref{sec:practical_implementation}, SCALE provides inference-time flexibility for controlling the sample quality by varying $\alpha$ and $\beta$. 
\cref{fig:ab_qualitative} visualizes their effects. 
%
Decreasing $\beta$ improves fidelity by amplifying the conditional signal, but stronger CFG pushes samples toward canonical templates and can change their semantic structure. Decreasing $\alpha$ instead targets fidelity directly while preserving semantic structure, though it may fall short when samples lie far from the data manifold (first row of \cref{fig:ab_qualitative}). More detailed visual analysis is provided in Appendix~\ref{app:visualization}.

\paragraph{Combination of SCG and CFG.}
Aforementioned analysis suggests that SCG and CFG are complementary to each other. Therefore, we combine SCG on top of the CFG prediction, which directly rectifies the CFG-generated samples toward the data manifold. This compensates for the limitation of each individual guidance: moderate CFG enhances prompt alignment and rectifies the underlying semantic structure, while SCG improves the sample quality on top of the CFG prediction. Utilizing CFG alone with a higher guidance scale may push samples away from the data manifold or incur reduced diversity as discussed in \cref{sec:motivation}. 
Notably, SCALE reduces to GFT~\cite{gft} at $\alpha=1$, which only approximates the CFG distribution. \cref{fig:ab_quantitative} shows that lowering $\alpha$ consistently improves FID and IS over this baseline. 
SCALE thus serves as a plug-and-play framework for improving guidance-free generation that maintains baseline performance rather than just adding complexity.


\paragraph{Training SCALE from scratch.}
Unlike most guidance-free methods that require a CFG-trained model~\cite{distillation,cca}, SCALE can be trained from scratch as it does not require a pretrained teacher model. Due to resource constraints, we verify this on DiT-B/2~\cite{dit}, where SCALE significantly improves guidance-free generation (\cref{tab:scratch_fid}). Scaling to larger models is left as future work.

\begin{table}[!t]
  \centering
  \small
  \setlength{\tabcolsep}{2pt}
  \begin{minipage}[t]{0.32\textwidth}
  \centering
  \captionof{table}{Guidance-free generation with trained from scratch.}
  \label{tab:scratch_fid}
  \begin{tabular}{llcc}
    \toprule
    \textbf{Model} & \textbf{Methods} & \textbf{FID} $\downarrow$ & \textbf{IS} $\uparrow$ \\
    \midrule
    \multirow{3}{*}{\shortstack{DiT-B/2 \\ \citep{dit}}}
    & Baseline & 44.8 & 30.7 \\
    & GFT~\citep{gft} & 9.04 & 166.6 \\
    & \cellcolor[RGB]{238, 238, 228}\textbf{{SCALE}} & \cellcolor[RGB]{238, 238, 228}\textbf{8.75} & \cellcolor[RGB]{238, 238, 228}\textbf{169.9} \\
    \bottomrule
    \end{tabular}
  \end{minipage}
  \hfill
  \begin{minipage}[t]{0.32\textwidth}
  \centering
  \captionof{table}{Ablation on interpolation type for SCALE.}
  \label{tab:ablation_interp_type}
  \begin{tabular}{llcc}
  \toprule
  \textbf{Model} & \textbf{Interp.} & \textbf{FID} $\downarrow$ & \textbf{IS} $\uparrow$ \\
  \midrule
  \multirow{3}{*}{\shortstack{DiT-B/2 \\ \citep{dit}}}
   & Baseline   & 44.8 & 30.7 \\
   & Linear & 8.67 & 171.8 \\
   & \cellcolor[RGB]{238, 238, 228}Patch & \cellcolor[RGB]{238, 238, 228}\textbf{8.60} & \cellcolor[RGB]{238, 238, 228}\textbf{172.4} \\
  \bottomrule
  \end{tabular}
  \end{minipage}
  \hfill
  \begin{minipage}[t]{0.32\textwidth}
  \centering
  \captionof{table}{Ablation on interpolation ratio $\lambda$.}
  \label{tab:ablation_interp_ratio}
  \begin{tabular}{llcc}
  \toprule
  \textbf{Model} & $\boldsymbol{\lambda}$ & \textbf{FID} $\downarrow$ & \textbf{IS} $\uparrow$ \\
  \midrule
  \multirow{3}{*}{\shortstack{LlamaGen-L \\ \citep{llamagen}}}
   & 1.0 & 2.59 & 245.9 \\
   & 0.5 & \textbf{2.58} & 245.2 \\
   & \cellcolor[RGB]{238, 238, 228}0.2 & \cellcolor[RGB]{238, 238, 228}\textbf{2.58} & \cellcolor[RGB]{238, 238, 228}\textbf{253.0} \\
  \bottomrule
  \end{tabular}
  \end{minipage}
  \end{table}
  
\paragraph{Interpolation type.}
As discussed in \cref{sec:practical_implementation}, we employ patch-wise mix-up to maintain compatibility with diverse generative architectures. \cref{tab:ablation_interp_type} also confirms that patch-wise mix-up provides robust improvements over linear interpolation by preserving local semantics of dense visual data.

\paragraph{Interpolation ratio.}
\cref{tab:ablation_interp_ratio} reports the impact of the interpolation ratio $\lambda$. For each $\lambda$, we report the best FID with its corresponding IS. Lower $\lambda$ achieves better FID and IS, indicating that a small but non-zero $\lambda$ is sufficient to provide stable and effective guidance.

\section{Conclusion}
We introduce \emph{SCALE}, a guidance-free training framework built on \emph{self-contrastive guidance} (SCG), a signal derived from the model's own predictions on real and self-generated samples.
Grounded in the information-based guidance theorem, SCG directly pushes the generation process toward the true data manifold.
Extensive experiments confirm that SCALE outperforms existing guidance-free baselines on both diffusion and AR models, and show complementary performance gain when combined with human preference-based fine-tuning.

\paragraph{Limitations.}
SCALE's nested guidance formulation introduces additional training overhead. Extensions to broader visual alignment tasks, post-training beyond DPO, and larger, more recent generative architectures remain open directions. See Appendix~\ref{app:limitations} for further discussion.

\section*{Acknowledgements}
This work was supported by Institute of Information \& Communications Technology Planning \& Evaluation(IITP) grant funded by the Korea government(MSIT) (RS-2024-00439020, Developing Sustainable, Real-Time Generative AI for Multimodal Interaction, SW Starlab).

{\small
\bibliographystyle{abbrvnat}
\bibliography{main}
}


\clearpage
\appendix

\newpage
\section{Derivation of Generalized Information-Based Guidance for Discrete Generative Models}
\label{app:gen_info_guidance}

In this section, we derive the optimization-perspective formulation of information-based guidance for discrete generative models in \eqref{eq:dsag_concise}, following the framework of Tang et al.~\cite{vqdiffusion} and Hur et al.~\cite{hur2024unlocking}.

Let $\Bar \vx_t$ denote a degraded sample that lacks some salient information $h_t$ present in the true sample $\vx_t$, such that the data distribution decomposes as $p(\vx_t) = p(\Bar \vx_t, h_t)$. 
Generalized information-based guidance recovers $\vx_t$ from $\Bar \vx_t$ by jointly maximizing the joint likelihood and amplifying the correlation between $\Bar \vx_t$ and $h_t$, captured by $p(h_t \mid \Bar \vx_t)$:
\begin{equation}
    \argmax_{\Bar \vx_t} \bigl[ \log p(\Bar \vx_t, h_t) + s \, \log p(h_t \mid \Bar \vx_t) \bigr],
    \label{eq:gen_obj}
\end{equation}
where $s \geq 0$ is the guidance scale that controls the strength of information amplification.

Applying the chain rule of probability $p(h_t \mid \Bar \vx_t) = p(\Bar \vx_t, h_t) / p(\Bar \vx_t)$, the conditional log-density expands as $\log p(h_t \mid \Bar \vx_t) = \log p(\Bar \vx_t, h_t) - \log p(\Bar \vx_t)$. Substituting into \eqref{eq:gen_obj} yields
\begin{equation}
    \argmax_{\Bar \vx_t} \bigl[ \log p(\Bar \vx_t, h_t) + s \bigl( \log p(\Bar \vx_t, h_t) - \log p(\Bar \vx_t) \bigr) \bigr],
    \label{eq:gen_obj_chain}
\end{equation}
which recovers \eqref{eq:dsag_concise} in the main text. This formulation explicitly amplifies the pointwise mutual information $\log p(\Bar \vx_t, h_t) - \log p(\Bar \vx_t) = \log p(h_t \mid \Bar \vx_t)$ by a factor of $s$ on top of the base log-density $\log p(\Bar \vx_t|h_t)= \log p(\vx_t)$.

\paragraph{Connection to Hur et al.~\cite{hur2024unlocking}.}
Hur et al.~\cite{hur2024unlocking} formulate the same objective with the conditional $p(\Bar \vx_t \mid h_t)$ in place of our joint $p(\Bar \vx_t, h_t)$. Since $\log p(\Bar \vx_t \mid h_t) = \log p(\Bar \vx_t, h_t) - \log p(h_t)$ and $\log p(h_t)$ does not depend on $\Bar \vx_t$, the two formulations differ only by an additive constant inside the $\argmax$ when $h_t$ is held fixed, and therefore lead to identical optima. We adopt the joint form because it avoids treating $h_t$ as a freely defined random variable, which the conditional $p(\Bar \vx_t \mid h_t)$ requires. Setting $p(\Bar \vx_t, h_t) = p(\vx_t)$ instead lets us work directly with the data distribution at hand without specifying a separate distribution over $h_t$.

\section{Proof of Theorem~\ref{prop:scale_optimality} and Analogous Result for AR Models}
\label{app:proof_optimality}

In this section, we derive the conditional mean SCG prediction at the diffusion regression optimum stated in \cref{prop:scale_optimality}, followed by the analogous result for AR models.

\paragraph{Proof for diffusion models.} Under the score matching objective, the optimal noise predictor $\vepsilon_{\theta^*}$ converges to the true scaled score from the noisy sample $\vx_t = \alpha_t \vx_0 + \sigma_t \vepsilon$:
\begin{equation}
    \vepsilon_{\theta^*}(\vx_t) = -\sigma_t \nabla_{\vx_t} \log p_t(\vx_t),
\end{equation}
where $t$ is the diffusion timestep.
Let $\tilde{\vx}_t$ be the self-generated sample from the model's current prediction. In the context of diffusion models as a noise predictor, $\tilde{\vx}_t$ can be obtained simply. Specifically, we define a mapping $g_\theta$ that first estimates the clean data using trained noise predictor $\vepsilon_\theta$ with Tweedie's formula~\cite{efron2011tweedie,kim2021noise2score} and subsequently re-noises it back to the same noise level $t$:
\begin{align}
    \tilde{\vx}_t &= g_\theta(\vx_t, \vepsilon)= \alpha_t \tilde{\vx}_0 + \sigma_t \vepsilon, \, \, \text{where}\\
    \tilde{\vx}_0 &= (\vx_t - \sigma_t \vepsilon_\theta(\vx_t))/\alpha_t, 
\end{align}

By the SCALE training objective, the trained model output is parameterized as the interpolation between the SCG branch and the self-generated branch:
\begin{equation}
    \vepsilon_\theta(\vx_t \mid \alpha) = \alpha\, \vepsilon^\text{scg}_\theta(\vx_t \mid \alpha) + (1-\alpha)\, \text{sg}\bigl[\vepsilon_\theta(\tilde \vx_t \mid 1)\bigr],
    \label{eq:scg_dm_sub}
\end{equation}
By construction, the $\alpha=1$ branch is the unconditional predictor,i.e., $\vepsilon_\theta(\cdot\mid1)\equiv\vepsilon_\theta(\cdot)$.
Hence, we may rewrite \eqref{eq:scg_dm_sub} as
\begin{equation}
    \vepsilon_\theta(\vx_t \mid \alpha)
    =
    \alpha\, \vepsilon^\text{scg}_\theta(\vx_t \mid \alpha)
    +
    (1-\alpha)\,
    \operatorname{sg}\bigl[
        \vepsilon_\theta(\tilde \vx_t)
    \bigr].
    \label{eq:scg_dm_sub_simple}
\end{equation}

The SCALE training directly optimizes $\vepsilon_\theta(\vx_t \mid \alpha)$ to minimize the score matching loss
\begin{equation}
\min_\theta \mathbb{E}_{(\vx_0,c) \sim p_{\text{data}}, \vepsilon, t } \left[ || \vepsilon - \vepsilon_\theta(\vx_t \mid \alpha) ||^2 \right].
\end{equation}

The reference branch is held fixed by stop-gradient. At an ideal regression optimum ($\theta=\theta^*$), the SCG branch therefore predicts the conditional mean of $(\vepsilon-(1-\alpha)\vepsilon_{\theta^*}(\tilde{\vx}_t))/\alpha$ given $\vx_t$. For fixed $t$ and $\alpha\in(0,1]$, using $\mathbb E[\vepsilon\mid\vx_t]=-\sigma_t\nabla_{\vx_t}\log p_t(\vx_t)$ in \eqref{eq:scg_dm_sub_simple} gives

\begin{equation}
    \alpha\, \vepsilon^\text{scg}_{\theta^*}(\vx_t \mid \alpha) + (1-\alpha)\, \mathbb E\bigl[\vepsilon_{\theta^*}(\tilde \vx_t)\mid\vx_t\bigr] = -\sigma_t \nabla_{\vx_t} \log p_t(\vx_t).
\end{equation}

Here $\tilde{\vx}_t=g_{\theta^*}(\vx_t,\vepsilon)$, and the expectation is over $\vepsilon\mid\vx_t$. By the assumed score accuracy at the reference inputs, $\vepsilon_{\theta^*}(\tilde \vx_t) = -\sigma_t \nabla_{\tilde \vx_t} \log p_t(\tilde \vx_t)$. Substituting this identity inside the expectation and solving for the SCG branch $f^\text{scg}_{\theta^*}(\vx_t \mid \alpha)$ gives

\begin{align}
    f^\text{scg}_{\theta^*}(\vx_t \mid \alpha) &= \frac{1}{\alpha} \Bigl[ -\sigma_t \nabla_{\vx_t} \log p_t(\vx_t) - (1-\alpha)\, \mathbb E\bigl[\vepsilon_{\theta^*}(\tilde \vx_t)\mid\vx_t\bigr] \Bigr] \\
    &= -\sigma_t \left[ \frac{1}{\alpha} \nabla_{\vx_t} \log p_t(\vx_t) - \left(\frac{1}{\alpha} - 1\right) \mathbb E\bigl[\nabla_{\tilde \vx_t} \log p_t(\tilde \vx_t)\mid\vx_t\bigr] \right]. 
    \label{eq:scg_exact2}
\end{align}

This form coincides with \eqref{eq:scale_optimal_form}, recovering the conditional mean SCG prediction. \qed

\paragraph{Proof for autoregressive models.}
We now derive the autoregressive (AR) counterpart, where the optimization directly works in the probability domain.
For AR models, $f_\theta$ is a {next-token predictor} that outputs a categorical distribution over the token vocabulary at each generation step.
To match the diffusion notation, we write $\vx_t$ for the next-token target at position $t$ given $\vx_{<t}$.

Concretely, the next-token logit is $\ell_\theta(\vx_t \mid \alpha)$ over the token vocabulary, and the predicted probability is $f_\theta(\vx_t \mid \alpha) = \mathrm{softmax}\bigl(\ell_\theta(\vx_t \mid \alpha)\bigr)$.
Under the cross-entropy objective in \eqref{eq:ar_loss}, the optimal next-token predictor converges to the true next-token conditional:
\begin{equation}
    f_{\theta^*}(\vx_t \mid \alpha) = p_t(\vx_t), \qquad \text{i.e., } \ell_{\theta^*}(\vx_t \mid \alpha) = \log p_t(\vx_t) + Z(\alpha),
\end{equation}
for all $\alpha$ in the training distribution, where $p_t(\vx_t)$ refers to the true next-token distribution at position $t$ given $\vx_{<t}$ and $Z(\alpha)$ is a vocabulary-wide additive constant absorbed by the softmax normalization.

By the SCALE training objective, the trained logit is parameterized as the interpolation between the SCG branch and the self-generated branch:
\begin{equation}
    \ell_\theta(\vx_t \mid \alpha) = \alpha\, \ell^\text{scg}_\theta(\vx_t \mid \alpha) + (1-\alpha)\, \text{sg}\bigl[\ell_\theta(\tilde \vx_t \mid 1)\bigr],
\end{equation}
where the self-generated token $\tilde \vx_t$ is drawn from the unconditional next-token distribution $f_\theta(\cdot \mid 1)$ via standard token-wise sampling.
At $\alpha = 1$, the SCG branch reduces to the unconditional next-token prediction, so $\ell_{\theta^*}(\tilde \vx_t \mid 1) = \log p_t(\tilde \vx_t) + Z(1)$ at convergence.

Substituting the convergence condition into the SCALE parameterization and solving for the SCG branch:
\begin{align}
    \alpha\, \ell^\text{scg}_{\theta^*}(\vx_t \mid \alpha) + (1-\alpha)\bigl[\log p_t(\tilde \vx_t) + Z(1)\bigr] = \log p_t(\vx_t) + Z(\alpha), \\
    \ell^\text{scg}_{\theta^*}(\vx_t \mid \alpha) = \frac{1}{\alpha} \log p_t(\vx_t) - \Bigl(\frac{1}{\alpha} - 1\Bigr) \log p_t(\tilde \vx_t) + Z',
    \label{eq:scg_ar_exact}
\end{align}
where $Z'$ denotes the sum of additive constants.
With $\alpha = 1/(1+s_\text{scg})$, applying softmax to \eqref{eq:scg_ar_exact} yields
\begin{equation}
    f^\text{scg}_{\theta^*}(\vx_t \mid \alpha) \;\propto\; p_t(\vx_t) \left[ \frac{p_t(\vx_t)}{p_t(\tilde \vx_t)} \right]^{s_\text{scg}},
    \label{eq:scg_ar_dist}
\end{equation}
which exactly matches the SCG-induced next-token distribution obtained by substituting $\Bar{\vx}_t = \tilde \vx_t$ and $h_t = \vx_t - \tilde \vx_t$ into the discrete information-based guidance objective in \eqref{eq:dsag_concise}. The SCALE-trained AR model therefore recovers the SCG distribution at convergence in the probability domain, mirroring the score-domain result for diffusion. \qed

\section{Analysis of the SCALE Training}
\label{app:scale_training_analysis}

The use of a self-generated reference in \eqref{eq:mug_gft} makes the SCALE training target dependent on the model's current predictions. We examine how this target evolves under stop-gradient updates and how prediction errors propagate through the PatchMix reference. The analysis is presented for diffusion models, where $f_\theta$ denotes the noise predictor and $\epsilon$ the ground-truth noise. We fix $t$ and $\alpha\in[\alpha_{\min},1]$ with $\alpha_{\min}>0$; conditional expectations average over $\epsilon\mid x_t$ and an independent PatchMix mask.

\subsection{Training with Stop-Gradient and PatchMix}
\label{app:scg_frozen_target}

\noindent
\paragraph{Across training iterations.}
Within each update, the SCALE training target uses a frozen reference. Under stop-gradient in \eqref{eq:mug_gft}, each training step follows the target-network iteration
\begin{equation}
\theta_{k+1} \leftarrow \theta_k - \mu \nabla_\theta \lVert \epsilon-\alpha \underbrace{f^{\text{scg}}_\theta(x_t \mid \alpha)}_{\text{optimization target}}-(1{-}\alpha)\underbrace{f_{\theta_k}(\tilde{x}_{t,\lambda} \mid 1)}_{\substack{\text{frozen reference}\\\text{(stop-gradient)}}}\rVert^2.
\label{eq:scg_target_update}
\end{equation}
Each iteration applies a gradient step to a regression loss with a frozen reference, evaluated at $\theta=\theta_k$. Both $\tilde{x}_{t,\lambda}$ and the prediction at that input are obtained from the $\alpha{=}1$ slice. At $\alpha=1$, the objective reduces to standard noise regression toward the fixed, model-independent target $f_{\theta^\ast}(x_t \mid 1)=-\sigma_t\nabla\log p_t(x_t)$. Within each update, the resulting reference dependency chain is acyclic (data $\to\alpha{=}1\to\alpha{<}1$); the target depends on the model but is anchored to the data distribution through the base slice's fixed noise-regression target. Since the slices share parameters, guided-slice updates can still change the base slice across iterations.

\noindent
\paragraph{Each gradient step.}
The reference evolves across successive model snapshots ($\theta_k\to\theta_{k+1}$), inducing a moving target for the $\alpha{<}1$ slices. The target variation is bounded by the base-slice change and vanishes as this change tends to zero. Rearranging the loss in \eqref{eq:scg_target_update} gives a sample target for the SCG branch. We denote its conditional mean given $x_t$ by $T[\theta_k]$, where $\theta_k$ specifies the fixed model snapshot used to construct and evaluate the reference:
\begin{equation}
T\lbrack\theta_k\rbrack(x_t,\alpha)=\frac{1}{\alpha}\bigl(-\sigma_t\nabla\log p_t(x_t)\bigr)-\bigl(\frac{1}{\alpha}{-}1\bigr)\mathbb E\bigl[f_{\theta_k}(\tilde{x}_{t,\lambda} \mid 1)\mid x_t\bigr],
\label{eq:scg_rebuttal_target}
\end{equation}
so a base-slice change $\Delta f:=f_{\theta_{k+1}}(\cdot\mid1)-f_{\theta_k}(\cdot\mid1)$ enters only through the last term. With a uniform change bound $\lVert\Delta f\rVert$ and a common Lipschitz constant $L$, we bound its effect through two contributions. First, the prediction at the reference input $\tilde{x}_{t,\lambda}$ itself changes, contributing $\frac{1-\alpha}{\alpha}\lVert\Delta f\rVert$. Second, the base prediction also changes the generated sample $\tilde{x}_t=x_t+\sigma_t(\epsilon-f_{\theta_k}(x_t\mid1))$, shifting the reference input $\tilde{x}_{t,\lambda}$. PatchMix reduces the displacement bound from $\sigma_t\lVert\Delta f\rVert$ to $\sqrt{\lambda}\,\sigma_t\lVert\Delta f\rVert$ in RMS. Feeding the shifted reference into the $L$-Lipschitz base predictor changes its output, giving a second contribution of at most $\frac{1-\alpha}{\alpha}\sqrt{\lambda}\,\sigma_t L\lVert\Delta f\rVert$. Together,
\begin{equation}
\lVert\Delta T\rVert\le\kappa\,\lVert\Delta f\rVert,\qquad \kappa=\frac{1-\alpha}{\alpha}(1+\sqrt{\lambda}\,\sigma_t L).
\label{eq:scg_rebuttal_kappa}
\end{equation}
If the base slice converges uniformly to $f_{\theta^\ast}(\cdot\mid1)$, then $\lVert\Delta f\rVert\to0$ and hence $\lVert\Delta T\rVert\to0$. The bound also gives $T[\theta_k]\to T[\theta^\ast]$, so the moving target settles to a fixed limit. The $\alpha{<}1$ slices converge to it if their regression error relative to $T[\theta_k]$ also vanishes.

\subsection{Optimum with PatchMix under Finite Capacity Bounds}
\label{app:scg_patchmix_bound}

At an ideal regression optimum ($\theta_k \rightarrow \theta^\ast$), the conditional mean of the mixed output matches the data score:
\begin{equation}
\alpha f^{\text{scg}}_{\theta^\ast}(x_t \mid \alpha)+(1{-}\alpha)\mathbb E\bigl[f_{\theta^\ast}(\tilde{x}_{t,\lambda} \mid 1)\mid x_t\bigr]=-\sigma_t\nabla\log p_t(x_t).
\label{eq:scg_patchmix_fixed_point}
\end{equation}
The regression argument in the proof of Theorem~\ref{prop:scale_optimality} does not depend on how the reference is constructed, so it also applies under PatchMix: substituting $f_{\theta^\ast}(\tilde{x}_{t,\lambda} \mid 1)=-\sigma_t\nabla\log p_t(\tilde{x}_{t,\lambda})$ assuming the base predictor matches the data score at the reference inputs and solving for the SCG branch gives
\begin{equation}
f^{\text{scg}}_{\theta^\ast}(x_t \mid \alpha)=-\sigma_t\lbrack\frac{1}{\alpha}\nabla\log p_t(x_t)-(\frac{1}{\alpha}{-}1)\mathbb E\bigl[\nabla\log p_t(\tilde{x}_{t,\lambda})\mid x_t\bigr]\rbrack.
\label{eq:scg_patchmix_rebuttal_optimum}
\end{equation}
Setting $\lambda{=}1$ recovers \eqref{eq:scale_optimal_form}, while $\lambda{=}0$ disables guidance at this ideal optimum. For a fixed real/generated pair, Bernoulli PatchMix gives
\begin{equation}
\mathbb{E}_{M_\lambda}\lVert \tilde{x}_{t,\lambda}-x_t\rVert^2=\lambda\lVert \tilde{x}_t-x_t\rVert^2,
\label{eq:scg_rebuttal_mask_distance}
\end{equation}
where $M_\lambda$ is the PatchMix mask in \eqref{eq:mixup}.

For finite capacity, let $\hat\theta$ be the trained model, $\delta_\alpha(x_t)$ the residual of its mixed output at conditioning $\alpha$ in \eqref{eq:scg_patchmix_fixed_point}, and $\delta(y):=f_{\hat\theta}(y \mid 1)-f_{\theta^\ast}(y \mid 1)$ the error of its $\alpha{=}1$ slice.
Since the generated reference depends on the model, write $\tilde{x}_{t,\lambda}$ and $\tilde{x}_{t,\lambda}^\ast$ for references constructed with $\hat\theta$ and $\theta^\ast$, respectively, using the same $\epsilon$ and mask. Using the mixed output $f_\theta(x_t\mid\alpha)$ from \eqref{eq:mug_gft}, the deviation from \eqref{eq:scg_patchmix_rebuttal_optimum} is
\begin{equation}
\begin{aligned}
&f^{\text{scg}}_{\hat\theta}(x_t\mid\alpha)-f^{\text{scg}}_{\theta^\ast}(x_t\mid\alpha)\\
&=\frac{1}{\alpha}\mathbb E\bigl[f_{\hat\theta}(x_t\mid\alpha)-f_{\theta^\ast}(x_t\mid\alpha)\mid x_t\bigr]-(\frac{1}{\alpha}{-}1)\mathbb E\bigl[f_{\hat\theta}(\tilde{x}_{t,\lambda}\mid1)-f_{\theta^\ast}(\tilde{x}_{t,\lambda}^\ast\mid1)\mid x_t\bigr]\\
&=\frac{1}{\alpha}\delta_\alpha(x_t)-(\frac{1}{\alpha}{-}1)\mathbb E\bigl[f_{\hat\theta}(\tilde{x}_{t,\lambda}\mid1)-f_{\theta^\ast}(\tilde{x}_{t,\lambda}\mid1)
+f_{\theta^\ast}(\tilde{x}_{t,\lambda}\mid1)-f_{\theta^\ast}(\tilde{x}_{t,\lambda}^\ast\mid1)\mid x_t\bigr]\\
&=\frac{1}{\alpha}\delta_\alpha(x_t)-(\frac{1}{\alpha}{-}1)\mathbb E\bigl[\delta(\tilde{x}_{t,\lambda})
+f_{\theta^\ast}(\tilde{x}_{t,\lambda}\mid1)
-f_{\theta^\ast}(\tilde{x}_{t,\lambda}^\ast\mid1)\mid x_t\bigr].
\end{aligned}
\label{eq:scg_rebuttal_deviation}
\end{equation}

Assume the prediction error is bounded by $\lVert \delta_\alpha(x_t)\rVert,\lVert \delta(x_t)\rVert\le e$ at the real input and $\lVert \delta(y)\rVert\le e+L_e\lVert y-x_t\rVert$ at the reference inputs. This assumption allows for reduced prediction accuracy as the reference input deviates from the real training data. Also assume $f_{\theta^\ast}(\cdot\mid1)$ is $L$-Lipschitz between $\tilde{x}_{t,\lambda}$ and $\tilde{x}_{t,\lambda}^\ast$. Applying these bounds and \eqref{eq:scg_rebuttal_mask_distance} to \eqref{eq:scg_rebuttal_deviation} gives
\begin{equation}
\begin{aligned}
&\lVert f^{\text{scg}}_{\hat\theta}(x_t\mid\alpha)-f^{\text{scg}}_{\theta^\ast}(x_t\mid\alpha)\rVert\\
&\le\frac{1}{\alpha}e+(\frac{1}{\alpha}{-}1)\bigl[e+\sqrt{\lambda}\bigl(L_e\mathbb E[\lVert\tilde{x}_t-x_t\rVert\mid x_t]+L\sigma_t e\bigr)\bigr]\\
&=(1+2s_{\text{scg}})\,e+s_{\text{scg}}\sqrt{\lambda}\bigl(L_e\mathbb E[\lVert\tilde{x}_t-x_t\rVert\mid x_t]+L\sigma_t e\bigr),
\end{aligned}
\label{eq:scg_patchmix_error_bound}
\end{equation}
where $\tilde{x}_t$ is the unmixed reference generated by $\hat\theta$ and $\alpha=1/(1+s_{\text{scg}})$. The $L\sigma_t e$ term accounts for the change in the reference input. At $\lambda{=}1$, a weak model can generate $\tilde{x}_t$ far from $x_t$, and the error at this reference enters the target amplified by $s_{\text{scg}}$. For a given model, $\lambda{<}1$ reduces the displacement-dependent terms in the bound by a factor of $\sqrt{\lambda}$.

Thus PatchMix brings the reference query $\tilde{x}_{t,\lambda}$ closer to the real input in RMS, tightening the assumed error bound on the base prediction used to approximate the optimum in \eqref{eq:scg_patchmix_rebuttal_optimum}. Under the stated assumptions, \eqref{eq:scg_patchmix_error_bound} shows that mixing degraded self-generated samples with real inputs suppresses the displacement-dependent contribution to the bound on the deviation from the PatchMix-SCG optimum. This supports the role of interpolation in \cref{sec:mixup} as a way to obtain a more reliable guidance anchor under limited model capacity.

\section{SCALE Training Pipeline for Autoregressive Models} \label{app:ar_scale}

We provide the training pipeline of SCALE for autoregressive (AR) models in \cref{alg:mug_ar}. The procedure mirrors the diffusion case (\cref{alg:mug_diff} in the main text), with the noise predictor $\vepsilon_\theta$ replaced by the next-token predictor $f_\theta$, and the score-matching loss replaced by the standard cross-entropy loss on the softmax-normalized logits. PatchMix is applied at the token level along the spatial dimension following \cref{eq:mixup}, and the nested guidance formulation in \eqref{eq:finalv2} is applied in the logit space before the softmax.

\begin{figure}[t]
    \centering
    \begin{minipage}{0.6\textwidth}
    \hrule height 0.8pt \vspace{3pt}
    \captionof{algorithm}{SCALE Training (AR)}
    \label{alg:mug_ar}
    \hrule \vspace{3pt}
    \begin{algorithmic}[1]
        \STATE Initialize next-token predictor $p_\theta$ and ratio $\lambda$
        \FOR{each gradient step}
        \STATE $\vx, c \sim p(\vx, c)$, $\alpha, \beta \sim U(0,1)$
        \STATE $\textcolor{blue}{\ell_\theta^\text{{scg-cfg}} = p_\theta(\vx|c,\alpha,\beta)}$ \hfill \small{\color{blue} \Comment{Target}}
        \STATE \small{\color{gray} \textit{// Additional to Training (Evaluation Mode)}}
        \STATE $\ell_\theta^u = p_\theta(\vx|\varnothing,1,1)$
        \STATE Sample $\tilde \vx$ from $\ell_\theta^u$
        \STATE $\tilde \vx_\lambda = \text{PatchMix}(\vx, \tilde \vx; \lambda)$ \hfill \Comment{\cref{eq:mixup}}
        \STATE $\ell_\theta^\lambda = p_\theta(\tilde \vx_\lambda|c,1,\beta)$
        \STATE \small{\color{gray} \textit{// Nested guidance formulation}} \hfill \Comment{\cref{eq:finalv2}}
        \STATE $\textcolor{blue}{\ell_\theta^\text{cfg} = \alpha \, \ell_\theta^\text{{scg-cfg}} + (1{-}\alpha) \,\text{sg}[\ell_\theta^\lambda]}$
        \STATE $\textcolor{blue}{\ell_\theta^c = \beta \, \ell_\theta^\text{cfg} + (1{-}\beta) \,\text{sg}[\ell_\theta^u]}$
        \STATE $\textcolor{blue}{p_\theta^c = \mathrm{softmax}(\ell_\theta^c)}$
        \STATE $\theta \leftarrow \theta - \mu \nabla_\theta \left[ - \log \textcolor{blue}{p_\theta^c} \right]$ \hfill \Comment{\cref{eq:ar_loss}}
        \ENDFOR
    \end{algorithmic}
    \vspace{2pt} \hrule height 0.8pt
    \end{minipage}
\end{figure}

\section{Further Implementation Details} \label{sec:detail}
In this section, we provide more detailed implementation details for clarity. As denoted in the main manuscript, we explicitly follow the training recipe from the base generative models and hyperparameters~\cite{gft}. For the fine-tuning, the training epoch is set to 1~5\% of the pre-training. 
We elaborate on the detailed training configurations used for SCALE training.

\paragraph{Input of $\alpha, \beta$.} Following implementation in GFT~\cite{gft}, we use Fourier embedding with MLP layers~\cite{ldm} to add conditioning for the models. For the fine-tuning, we adopt zero initialization for the conditioning network~\cite{controlnet}. With this formulation, the additional embeddings for the $\alpha$ and $\beta$ provide a lightweight (0.0012\%) parameter overhead, and negligible impact on inference latency.

\paragraph{PatchMix patch size.}
We set the PatchMix patch size in \eqref{eq:mixup} to align with the model's self-attention granularity, so that each mixed patch respects the unit at which the model processes spatial information. For Transformer-based models with an explicit patch tokenizer, this is the token/patch size of the input: 2 for DiT-XL/2 and DiT-B/2~\cite{dit} (latent patches of size 2), and 1 for LlamaGen-L~\cite{llamagen} (one VQ token per spatial location). For UNet-based models such as SD\,1.5~\cite{ldm}, which contain multiple self-attention resolutions, we use the finest self-attention resolution in the latent space, corresponding to a patch size of 2.

\paragraph{Training and sampling.}
For fine-tuning, we limit training to less than $5\%$ of the pre-training budget and inherit all hyperparameters from the base model. For training from scratch, we follow the original training recipe of the base model. For diffusion sampling, we use 50-step DPM-Solver++~\cite{lu2022dpmpp}. For text-to-image, we additionally use LoRA~\cite{ryu2023low} due to compute constraints (full setup in the SD\,1.5 paragraph below).
In our experiments, we found that combining CFG with SCG yields the best performance, as it provides an orthogonal guidance direction (see \ref{sec:analysis}). To ensure that the model focuses on the effective range of guidance strengths, we set a minimum $\alpha$ of 0.5 for class-conditional models and 0.25 for text-to-image models during training, which corresponds to setting the maximum SCG strength as 1.0 and 3.0, respectively.

\subsection{Fine-tuning}
\paragraph{LlamaGen}
We utilize LlamaGen-L for fine-tuning AR models with SCALE. We fine-tune these models with 15 epochs with a batch size of 256. The learning rate is set to 2e-4 with cosine decay. All pre-trained models are obtained from the official repository~\cite {llamagen}. We obtain $\tilde \vx_t$ by categorical sampling from the unconditional logits, without temperature scaling or top-$k$ truncation, so that $\tilde \vx_t$ is drawn directly from the model's distribution.

\paragraph{DiT}
For the main experiments and ablation studies, we fine-tune DiT-B/2 and DiT-XL/2 with SCALE using a batch size of 256 and a learning rate of 1e-4 with cosine decay. The EMA decay rate is set to 0.9999. 
We observed that initializing from a pretrained from GFT objective model yields slightly better performance. We posit that incorporating self-contrastive guidance subsequent to CFG training is advantageous.
%
For sampling, we adopt 50-step DPM-Solver++~\cite{lu2022dpmpp}.


\paragraph{Stable Diffusion v1.5 (Text-to-Image)}
For text-to-image generation at $512\times512$ resolution, we fine-tune Stable Diffusion v1.5 (SD\,1.5)~\cite{ldm} with SCALE on LAION-Aesthetics 5+~\cite{schuhmann2022laion}, which contains 18M text-image pairs. Due to compute constraints, we use LoRA~\cite{hu2022lora} with rank $r=128$ and scaling factor $128$, trained for 50k iterations using a batch size of 256, a learning rate of $1\text{e}{-5}$, and 1000 warmup steps. We initialize from a GFT fine-tuned SD\,1.5 checkpoint and apply SCALE on top, which compensates for the limited capacity of LoRA. For evaluation, zero-shot FID-30k, CLIP Score~\cite{radford2021learning}, and Aesthetic Score~\cite{schuhmann2022laion} are measured on 30k images generated from COCO 2014~\cite{lin2014microsoft} validation prompts, while the human-preference metrics including PickScore~\cite{pickscore}, HPSv2.1~\cite{hpsv2}, and Image Reward~\cite{imagereward} are measured on the Pick-a-Pic test prompts~\cite{pickscore}. For all quantitative and qualitative results, we utilize 50-step DPM-Solver++~\cite{lu2022dpmpp} for sampling.

\paragraph{DiffusionDPO on top of SCALE.}
For the DPO experiment, we fine-tune the GFT and SCALE-trained SD\,1.5 with DiffusionDPO~\cite{wallace2024diffusion} following the \texttt{diffusers} implementation\footnote{\href{https://github.com/huggingface/diffusers/blob/main/examples/research_projects/diffusion_dpo}{\texttt{github.com/huggingface/diffusers/research\_projects/diffusion\_dpo}}}. We use LoRA with rank $r=8$ and $\alpha=8$, learning rate $1\mathrm{e}{-5}$, gradient accumulation of $8$ for an effective batch size of $128$, and train for 10k iterations on a single RTX 4090 GPU using the Pick-a-Pic dataset~\cite{pickscore}. 

During DPO training, we fix the SCALE conditioning at $\alpha=1, \beta=1$ (i.e., the unamplified branch) and only vary $\alpha,\beta$ at inference to control the SCG and CFG strengths. This mirrors the standard practice in language-model alignment, where DPO training is performed with sampling temperature $1.0$ and the temperature is lowered only at inference~\cite{rafailov2023direct,wallace2024diffusion}. Training GFT with DPO follows the same practice, setting $\beta=1$ for the DPO training and a lower $\beta$ at inference, keeping other training hyperparameters unchanged, such as LoRA rank, learning rate, and effective batch size.

\subsection{Training From Scratch }
Due to the computational resource limitations within an academic research setting, pre-training with large-scale visual generative models was not feasible. In this paper, to demonstrate SCALE's applicability to pre-training, we primarily conduct experiments using DiT-B/2 models~\cite{dit}. This model was selected as it requires moderate training resources, requiring 80 epochs (400k iterations with a batch size of 256) for training and taking $\sim3$ days with 8 RTX4090 GPUs. For the pre-training, we drop conditions for the generated sample branch of SCALE.
In this case, $\Bar \vx_t=\tilde \vx_t$ and the $h_t$ is a combination of $\vx_t-\tilde \vx_t$ and the condition $c$ for the information-based guidance formulation in Eqs.~(\ref{eq:sag_concise}) and (\ref{eq:dsag_concise}). 
By omitting the conditional signal, the objective initially emulates a CFG, which effectively stabilizes the model before the self-contrastive signal becomes reliable. This approach facilitates a natural curriculum, where the model first learns CFG signals and seamlessly transitions to incorporating self-contrastive signals as generation quality matures.

\subsection{Human Preference Study} \label{app:human_study}

\begin{figure}[h]
    \centering
    \includegraphics[width=0.85\linewidth]{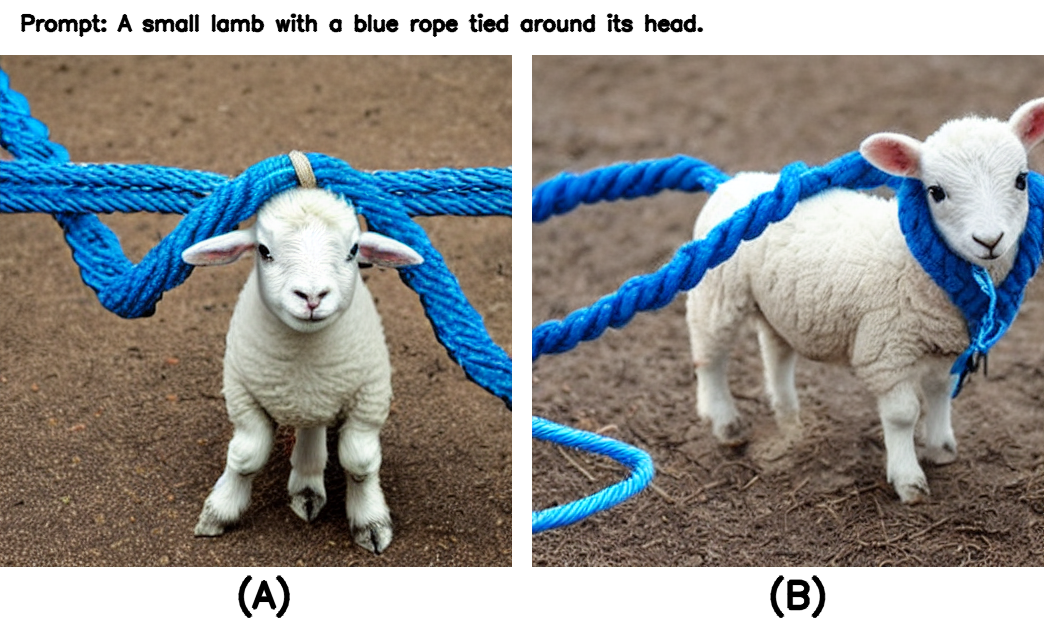}
    \caption{Example trial from the blind A/B human preference study. \textbf{Left (A):} GFT~\cite{gft}. \textbf{Right (B):} SCALE. Participants selected the image without these labels.}
    \label{fig:human_study_ui}
\end{figure}

We complement the automated human preference metrics on SD\,1.5 with a user study that directly compares SCALE against the strongest guidance-free baseline (GFT) in a side-by-side setting.
The study is conducted as a \textbf{blind A/B test} with \textbf{20 participants}, each responding to \textbf{15 image pairs} for a total of 300 votes. For each trial, the participant is shown one text prompt and two generated images (A and B), one from GFT and one from SCALE, with model identities hidden and the left/right order randomized per trial. The participant is asked to select the image they prefer overall, or to choose ``Tie'' if the two are of comparable quality. Prompts are sampled from the Pick-a-Pic test set so that the comparison covers diverse content. An example trial is shown in \cref{fig:human_study_ui}.

Following common practice in text-to-image preference studies, participants are asked to consider three axes when making their choice:
\begin{itemize}[leftmargin=2em,topsep=2pt,itemsep=2pt]
    \item \textit{Text alignment}: which image more accurately represents the content of the given prompt;
    \item \textit{Visual quality}: which image exhibits better clarity, texture, and fine details;
    \item \textit{Photorealism}: which image looks more realistic and aesthetically pleasing, without artificial artifacts.
\end{itemize}



\section{Related Works} \label{app:related}

\subsection{Visual Generative Models}
Likelihood-based visual generative models have been actively researched along several architectural lines. Diffusion models~\citep{ho2020denoising,song2020score,nichol2021improved,ldm,dit,glide} learn the data score and generate samples through iterative denoising. Autoregressive models~\cite{vqgan,llamagen,var,dalle2,tang2024hart,li2024autoregressive,han2025infinity,yu2025randomized} factorize the joint distribution into a sequence of next-token predictions over discrete or continuous tokens. Masked generative models~\cite{maskgit,li2023mage,chang2023muse,weber2024maskbit} predict masked tokens in parallel and trade autoregressive ordering for faster sampling. Flow matching models~\cite{liu2022flow,lipman2022flow,esser2024scaling,flux2024} learn a velocity field between noise and data and have recently scaled to large text-to-image systems such as Stable Diffusion~3~\cite{esser2024scaling} and FLUX~\cite{flux2024}. Despite their architectural differences, all of these families produce samples whose unconditional fidelity falls short of the training distribution and rely on additional guidance signals to close the gap.

\subsection{Contrastive Signals for High-Fidelity Visual Generation}
A broad range of methods enhances sample fidelity by leveraging a contrastive signal that distinguishes desired samples from undesired ones.

\paragraph{Information-based guidance.}
Classifier-free guidance (CFG)~\cite{ho2022classifier} contrasts a conditional and an unconditional prediction at sampling time, amplifying the conditional signal.
Information-based guidance~\cite{sag,hur2024unlocking} generalizes CFG in terms of model-based interpretation, constructing the undesired sample from a degraded version of the input itself. From this perspective, CFG is a special case where the degradation is dropping the condition. Formally, these methods amplify any salient information $h_t$ relative to a degraded reference $\bar\vx_t$.
Self-Attention Guidance (SAG)~\cite{sag} utilized a Gaussian blur to amplify high-frequency details of the input, with a self-attention mask to stabilize the guidance. Self-Guidance~\cite{hur2024unlocking} utilizes an auxiliary model to produce a degraded reference input for the semantic vector-quantized tokens~\cite{vqvae2,vqgan}. While effective, these methods require additional forward passes and rely on heuristics for constructing degraded inputs, which are often domain-specific and hard to generalize to other domains. 

\paragraph{Weak-model guidance.}
These methods leverage a weaker version of the model itself to produce undesired samples.
These methods generalize CFG in terms of model-based interpretation by replacing the unconditional reference with a weaker version of the model itself: autoguidance~\cite{autoguidance} contrasts the main model with a smaller or earlier checkpoint, Perturbed Attention Guidance~\cite{pag} and Smoothed Energy Guidance~\cite{hong2024smoothed} construct weak references through attention perturbation, and  $S^2$-guidance~\cite{s2guidance} contrasts with random subnetworks of the main model, constructed by stochastic block-dropping.

\paragraph{Guidance-free generation.}
Guidance-free methods incorporate CFG during training to enable sampling with a single model evaluation per step.
Guidance distillation~\citep{distillation} trains a student to match the guided predictions of a pretrained diffusion teacher.
Condition Contrastive Alignment (CCA)~\citep{cca} fine-tunes AR models using matched and mismatched image--condition pairs and a frozen pretrained reference model, with a contrastive alignment objective whose optimum recovers the CFG sampling distribution.
Guidance-Free Training (GFT)~\citep{gft} retains the standard likelihood objective and reparameterizes the conditional model output as an interpolation between the sampling and unconditional predictions. This enables guidance-free training from scratch across diffusion, AR, and masked generative models.
Model-guidance (MG)~\citep{tang2025diffusion} similarly learns CFG during training, but adds a stop-gradient CFG residual directly to the regression target rather than reparameterizing the model output.
These approaches derive their guidance signal from the conditional--unconditional contrast of CFG. SCALE introduces a complementary contrast between real and self-generated samples and learns it alongside CFG, aiming to improve generation beyond reproducing CFG behavior.

\paragraph{Reinforcement learning-based methods.}
Developed from the language model alignment literature, reinforcement learning (RL) post-training methods often construct a desired and undesired contrast through reward or preference signals.
Direct preference optimization (DPO)~\cite{rafailov2023direct,wallace2024diffusion,zhu2025dspo,yang2024using,liang2024step,deng2024flow} optimizes the policy directly on paired win--lose paired preference data. In this case, the model is to maximize the win probability over the lose probability, which implicitly encourages the model to generate samples that are preferred by the human.
Recently, group relative policy optimization (GRPO)~\cite{shao2024deepseekmath,liu2025flow,xue2025dancegrpo,he2025tempflow} samples a group of responses for each prompt and uses their reward statistics within the group as a critic-free baseline, optimizing the model with the resulting relative advantages.
While effective, these methods require an external network or curated preference data and typically still rely on CFG at inference for generating high-quality samples in the visual domain.


\section{Additional Results}


\subsection{Computational Overhead Analysis} \label{app:computational}

This section details the experimental setup and per-row analysis of \cref{tab:comparison} in the main text.

\paragraph{Measurement setup.}
For diffusion models, we report the costs of fine-tuning from a pretrained checkpoint on $8\times$RTX 4090 GPUs. We use DiT-B/2~\cite{dit} as the base model rather than DiT-XL/2 because guidance distillation requires loading the teacher network alongside the student, and the resulting peak memory exceeds the 24 GB capacity of a single RTX 4090 at the DiT-XL/2 scale. DiT-B/2 fits comfortably for all methods and yields a fair comparison under identical hardware. For AR models, the numbers for CCA~\cite{cca} and GFT are taken directly from the GFT paper~\cite{gft}, which measured costs on VAR-d30~\cite{var}.

\paragraph{Per-row analysis.}
We unpack each row of \cref{tab:comparison} below.

\textit{Train time / step.}
The relative cost is determined by the number of model evaluations per gradient step.
Guidance distillation~\cite{distillation} evaluates the teacher's conditional and unconditional predictions in addition to the student's forward pass, yielding a $\times 1.12$ overhead.
GFT performs only one extra unconditional forward as a stop-gradient reference, which gives the lightest overhead at $\times 1.06$.
SCALE matches Distillation's $\times 1.12$ because, in addition to the conditional forward, it computes a stop-gradient unconditional reference and a stop-gradient SCG reference output, both without backward passes.
CCA on AR models reports a higher $\times 1.69$ as it requires sampling negative tokens and an additional contrastive forward per step.

\textit{Params.}
SCALE introduces only the conditioning embeddings for $\alpha$ and $\beta$ (Fourier features followed by a small MLP), accounting for roughly $0.0012\%$ of the base model parameters. The added parameters are therefore negligible for both model size and inference latency, leaving the parameter ratio at $\times 1.00$.

\textit{VRAM.}
Guidance distillation must keep the teacher network resident in memory alongside the student, which inflates VRAM to $\times 1.21$ on DiT-B/2 (and prevents fitting DiT-XL/2 on a 24 GB GPU). CCA similarly requires storing reference statistics for the contrastive objective, leading to $\times 1.39$ on AR. GFT and SCALE, in contrast, share the same single network for all branches and only run extra forward passes without gradient retention, so VRAM remains at $\times 1.00$.

\textit{Inference.}
All single-network guidance-free methods (Distill, CCA, GFT, SCALE) perform a single forward pass per sampling step rather than CFG's two, halving the inference cost to $\times 0.50$.

\paragraph{Summary.}
SCALE matches GFT in memory and inference cost while keeping training-time overhead within the same order as guidance distillation, despite delivering a strictly richer training target ($p_t^\text{cfg}(\vx_t|c)/p_t^\text{cfg}(\tilde\vx_t|c)$ vs. $p_t^\text{cfg}(\vx_t|c)$). The added cost is concentrated in two gradient-free reference passes during training and vanishes entirely at sampling time.

\subsection{Visualization of PatchMix Interpolation} \label{app:mixup_visualize}

\begin{figure}[!t]
    \centering
    \includegraphics[width=\linewidth]{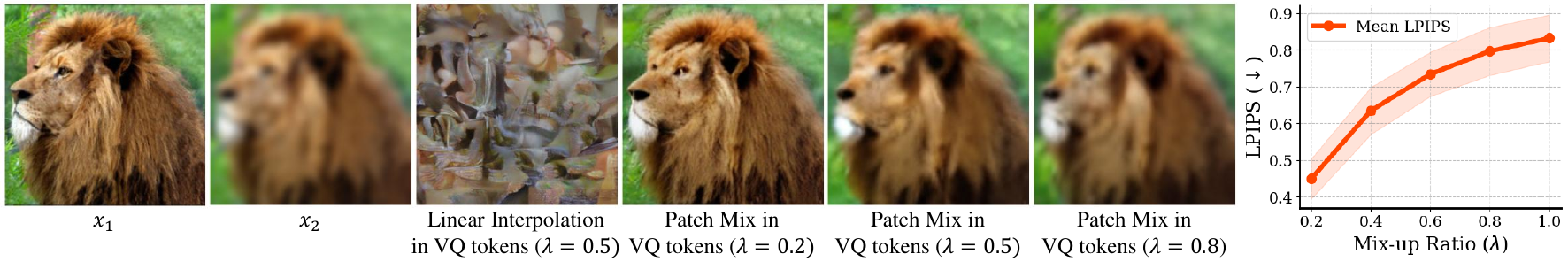}
    \caption{\textbf{Effect of interpolation choice.}
    \textbf{(Left)} Linear vs. patch-wise interpolation in the VQ token space (LlamaGen tokenizer~\cite{llamagen}). Linear interpolation collapses to noise at intermediate $\lambda$, while PatchMix transitions smoothly between $\vx_1$ and $\vx_2$.
    \textbf{(Right)} LPIPS distance between real samples and PatchMix-interpolated samples on DiT-XL/2~\cite{dit}: lower $\lambda$ keeps the interpolated sample closer to the real-data manifold, providing a stable anchor for self-contrastive guidance.}
    \label{fig:mixup_visualize}
\end{figure}

\cref{fig:mixup_visualize} provides additional support for the patch-wise design choice in \cref{eq:mixup}.
\textbf{(Left)} We compare linear and patch-wise interpolation in the vector-quantized (VQ) token space using the LlamaGen tokenizer~\cite{llamagen}. For linear interpolation, we apply $\lambda \vx_1 + (1-\lambda)\vx_2$ on the token indices and round to the nearest integer. Despite the high structural similarity between the two source images, linear interpolation fails to preserve any semantic content at intermediate $\lambda$, since rounded VQ indices do not lie on a smooth manifold of the codebook. PatchMix, by replacing whole patches rather than interpolating coordinates, transitions smoothly from $\vx_1$ to $\vx_2$ as $\lambda$ increases from $0$ to $1$ and remains semantically interpretable throughout.
\textbf{(Right)} We measure the LPIPS distance~\cite{zhang2018unreasonable} between $50$ real images and their PatchMix-interpolated counterparts $\tilde\vx_{t,\lambda}$ on DiT-XL/2~\cite{dit}, varying $\lambda$. The perceptual distance decreases monotonically as $\lambda$ shrinks, confirming that a smaller $\lambda$ keeps $\tilde\vx_{t,\lambda}$ close to the real-data manifold and therefore provides a stable anchor for the self-contrastive guidance.

\subsection{Comparison with Guided Sampling Methods} \label{app:guided_ddpm}

\cref{tab:guided_ddpm} reports an additional comparison under the original 250-step DDPM solver of DiT-XL/2~\cite{dit}, where the reported numbers for SAG~\cite{sag} and G++~\cite{discg} are taken from their original papers under their original solvers, ensuring a faithful baseline. Even without any guidance at sampling time, SCALE attains the best IS and a competitive FID, surpassing CFG and remaining on par with guided sampling methods that incur substantial extra inference cost.

While SAG and G++ further improve over CFG, they do so at the price of an additional forward pass per step and rely on settings that limit their generalizability. SAG operates at a tiny guidance scale of $0.005$, providing only a marginal guidance signal that is unlikely to significantly improve quality over CFG. G++ adopts a timestep-aware guidance schedule tightly coupled to its solver: when transferred from the 250-step DDPM solver to the 50-step DPM-Solver++, we observe a significant degradation in IS (compared to \cref{tab:guided} in the main text), indicating that its gains are sensitive to the sampling configuration. Moreover, both methods are specifically designed for score-based diffusion models and cannot be readily applied to discrete generative models such as AR models.

In contrast, SCALE delivers consistent improvements without any guidance at sampling time, achieving the highest IS in this comparison. Crucially, SCALE is agnostic to the choice of diffusion solver, applies uniformly to both diffusion and AR models, and requires no model-specific tuning, making it a broadly applicable framework for high-fidelity guidance-free generation.

\begin{table}[!t]
    \caption{Comparison with guided sampling methods on the ImageNet generation at $256\times256$.}
    \label{tab:guided_ddpm}
    \centering
    \small
    \setlength{\tabcolsep}{4pt}
    \begin{tabular}{lcccccc}
    \toprule
    \textbf{Methods} & \textbf{Guid.} & \textbf{NFEs} $\downarrow$ & \textbf{FID} $\downarrow$ & \textbf{IS} $\uparrow$ & \textbf{Precision} $\uparrow$ & \textbf{Recall} $\uparrow$ \\
    \midrule
    DiT-XL/2~~\citep{dit}      & \textcolor{red}{\ding{55}} & 250 & 9.62 & 121.5 & 0.67 & \textbf{0.67} \\
    $\quad+$CFG ~\citep{ho2022classifier} & \textcolor{green!60!black}{\checkmark} & 250 $\times$ 2 & 2.27 & \underline{278.24} & \textbf{0.83} & 0.57 \\
    $\quad+$SAG ~\citep{sag} & \textcolor{green!60!black}{\checkmark} & 250 $\times$ 3 & 2.11 & -- & -- & -- \\
    $\quad+$G++ ~\citep{discg} & \textcolor{green!60!black}{\checkmark} & 250 $\times$ 3 & \textbf{1.83} & {281.53} & 0.78 & \underline{0.64} \\
    \rowcolor[RGB]{230,230,255}
    $\quad+${{SCALE} (Ours)} & \textcolor{red}{\ding{55}} & 250 & \underline{2.10} & \textbf{288.08} & \underline{0.81} & 0.59 \\
    \bottomrule
    \end{tabular}
\end{table}

\section{Additional Visualizations and Detailed Analysis} \label{app:visualization}

\begin{figure}[!t]
    \centering
    \includegraphics[width=\linewidth]{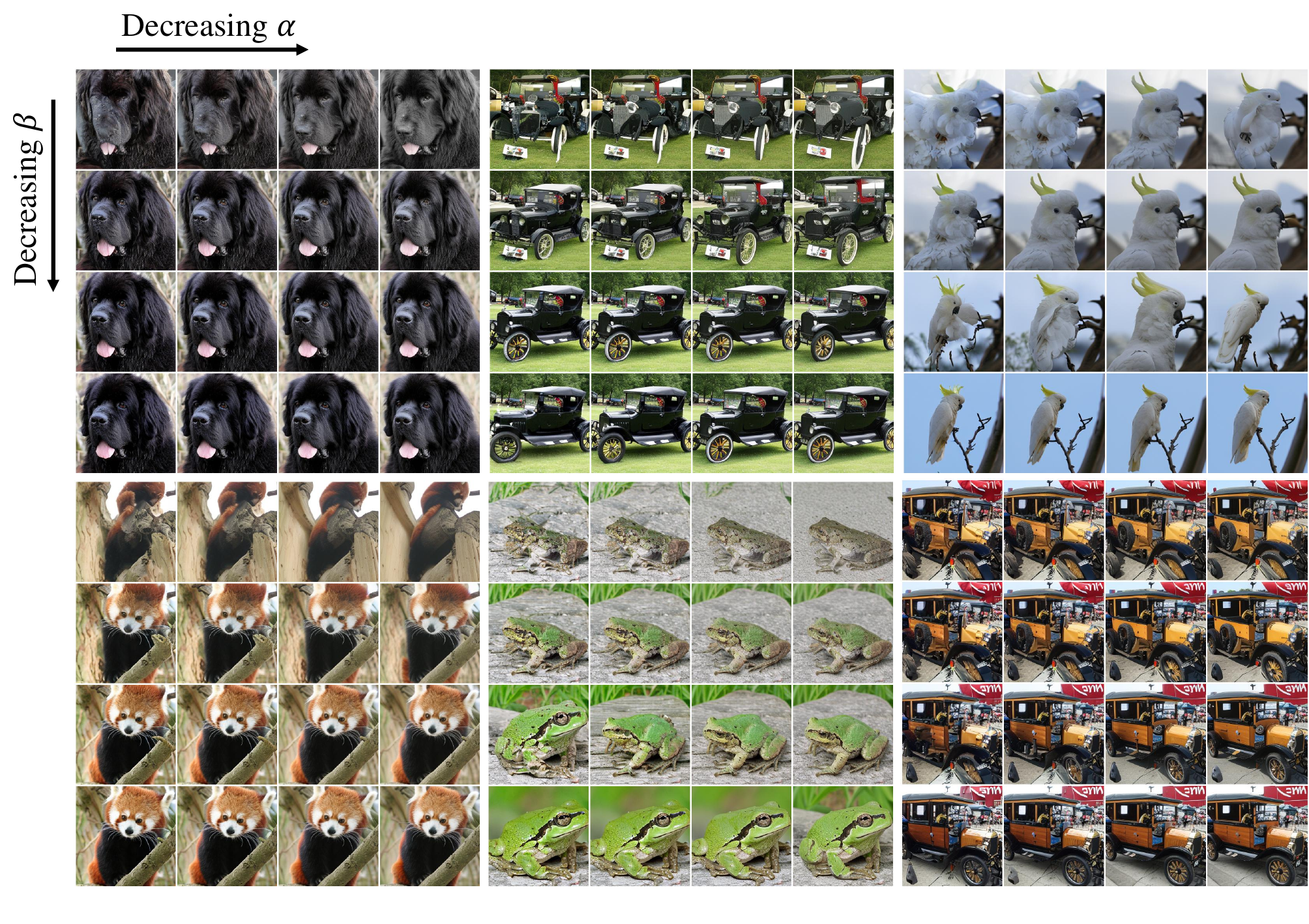}
    \caption{Visualization of SCALE by varying $\alpha$ and $\beta$ using DiT-XL/2~\cite{dit}.}
    \label{fig:visualization_ditxl}
\end{figure}

\begin{figure}[!t]
    \centering
    \includegraphics[width=\linewidth]{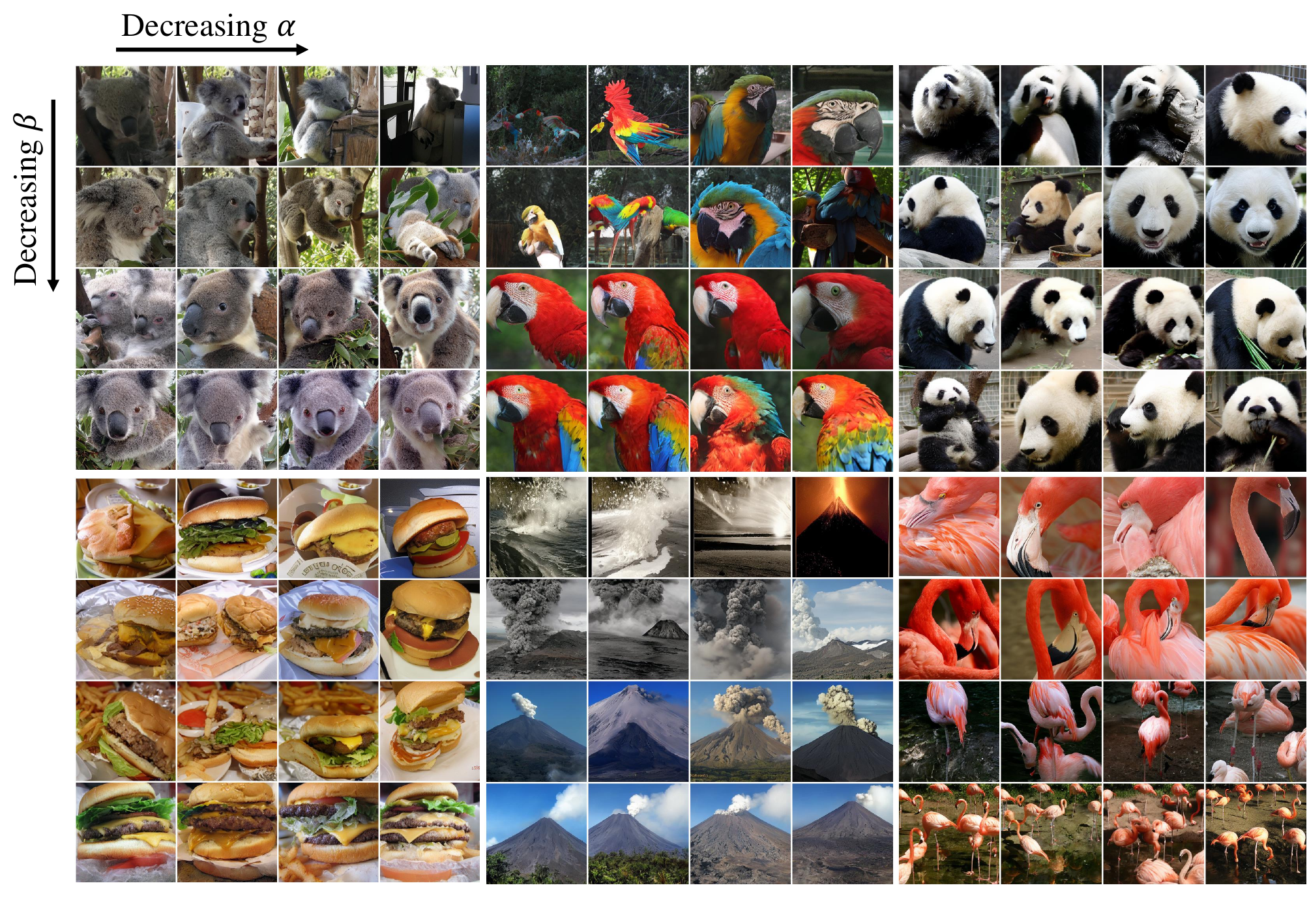}
    \caption{Visualization of SCALE by varying $\alpha$ and $\beta$ using LlamaGen-L~\cite{llamagen}.}
    \label{fig:visualization_llamagen}
    \vspace{5mm}
\end{figure}

\cref{fig:visualization_ditxl} and \cref{fig:visualization_llamagen} show the samples of SCALE by varying $\alpha$ and $\beta$ using DiT-XL/2~\cite{dit} and LlamaGen-L~\cite{llamagen}, respectively. We note that \textit{all samples are generated with a single model without guidance at sampling time}, and varying $\alpha$ and $\beta$ at inference time does not incur additional computational cost.

\paragraph{Effect of varying $\beta$ (implicit CFG).}
By decreasing $\beta$, the sample quality is improved by amplifying the conditional signal. However, it often pushes samples toward canonical templates and the semantic structure of samples can be steeply changed as CFG becomes stronger, as shown in each column of the image grid in \cref{fig:visualization_ditxl} and \cref{fig:visualization_llamagen}. Consequently, strong CFG (i.e., lower $\beta$) produces samples with reduced diversity as shown in the bottom row of each image grid.

\paragraph{Effect of varying $\alpha$ (implicit SCG).}
By decreasing $\alpha$, the sample quality is improved by amplifying the self-contrastive signal. It also improves the sample quality, but preserves the semantic structure as shown in each row of the image grid in \cref{fig:visualization_ditxl} and \cref{fig:visualization_llamagen}.
As a result, strong SCG (i.e., lower $\alpha$) produces more diverse high-quality samples as shown in the rightmost column of each image grid. However, limitations exist. Since SCG improves the sample quality while preserving the semantic structure, it may not be able to produce high-quality samples if the unguided sample is far from the data manifold, as shown in the top row of each image grid. 

\paragraph{Combination of SCG and CFG.}
Aforementioned analysis suggests that SCG and CFG are complementary to each other. Therefore, we combine SCG on top of the CFG prediction, which directly rectifies the CFG-generated samples to the data manifold. This compensates for the limitation of each individual guidance: moderate CFG enhances prompt alignment and rectifies the underlying semantic structure, while SCG improves the sample quality on top of the CFG prediction. Utilizing CFG alone with a higher guidance scale may push samples away from the data manifold or incur reduced diversity as discussed in \cref{sec:motivation}.

\begin{figure}[!t]
    \centering
    \includegraphics[width=\linewidth]{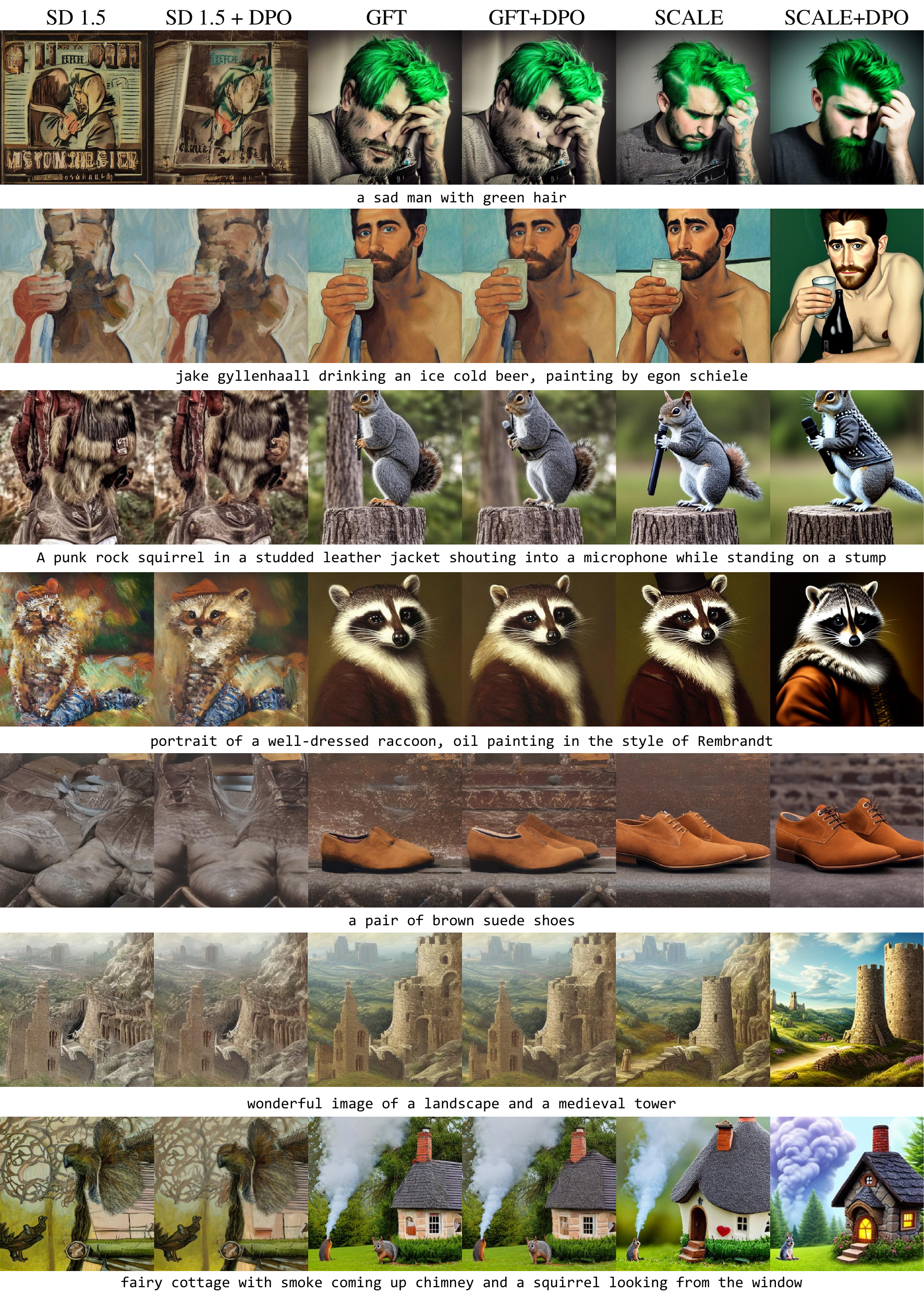}
    \caption{Guidance-free generation results of base SD 1.5~\cite{ldm}, GFT, and SCALE (ours), with and without DPO post-training~\cite{wallace2024diffusion}. Prompts are sampled from the Pick-a-Pic test set~\cite{pickscore} and PartiPrompts~\cite{partiprompts}.}
    \label{fig:visualization_dpo}
    \vspace{3mm}
\end{figure}

\begin{figure}[!h]
    \centering
    \includegraphics[width=\linewidth]{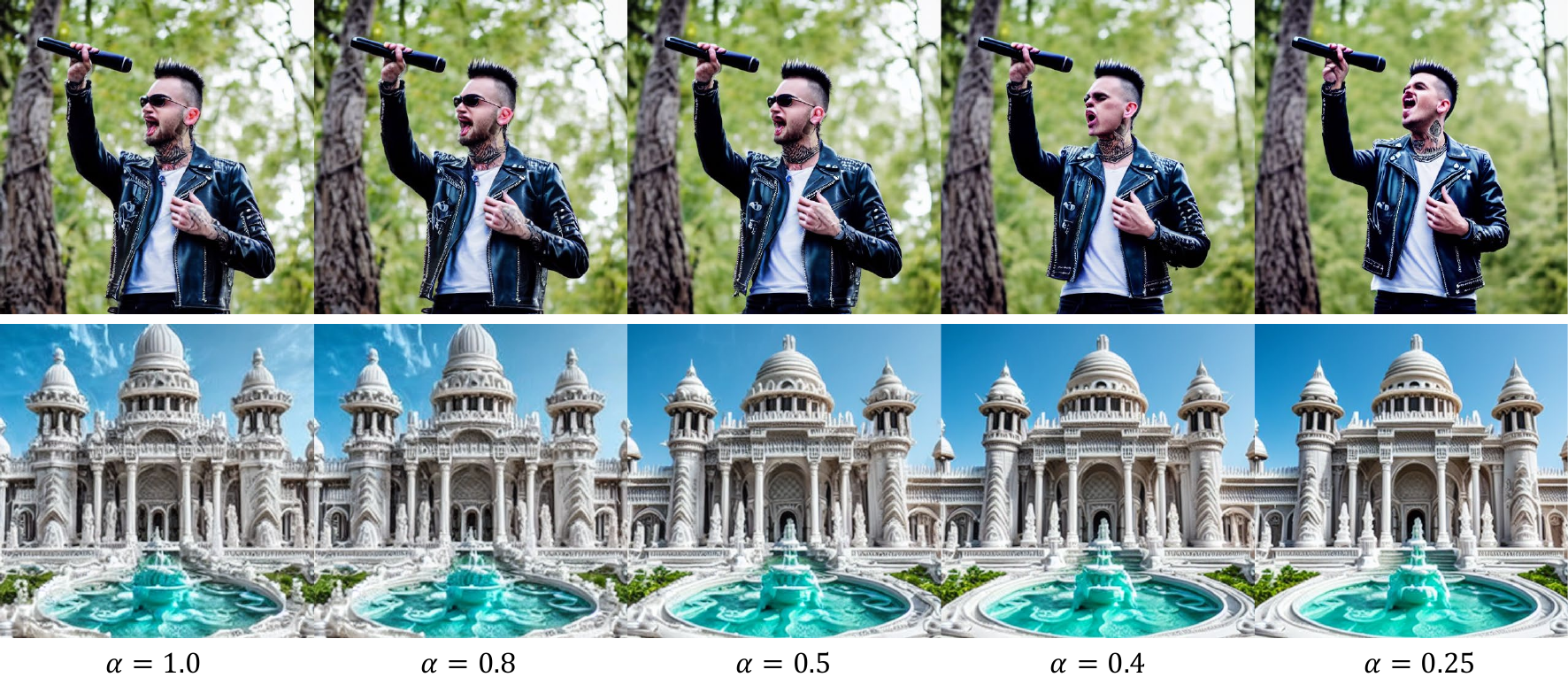}
    \caption{Visualization of SCALE by varying $\alpha$ using SD 1.5~\cite{ldm} with DPO post-training~\cite{wallace2024diffusion}.}
    \label{fig:visualization_dpo_alpha_sweep}
\end{figure}

\paragraph{Visualization with DPO post-training.}
\cref{fig:visualization_dpo} provides additional visualization of guidance-free generation with DPO post-training using stable diffusion v1.5 (SD 1.5)~\cite{ldm}. We note that \textit{all samples are generated with a single model without guidance at sampling time}, and varying $\alpha$ and $\beta$ does not incur additional computational cost. DPO post-training without CFG at sampling time fails to produce high-quality samples. While GFT and its DPO fine-tuned model improve the sample quality, SCALE further enhances the sample quality by directly refining the samples towards the data manifold, implicitly. \cref{fig:visualization_dpo_alpha_sweep} also confirms that amplifying the SCG signal by decreasing $\alpha$ gradually improves the sample quality of the DPO fine-tuned SCALE model.

\section{Limitations and Future Work} \label{app:limitations}

While SCALE provides consistent gains across both class-conditional and text-to-image generation on diffusion and AR architectures, several questions remain open.

\paragraph{Unified guidance formulation.}
The current SCALE framework is built upon the hierarchical guidance formulation, which combines the CFG and the SCG. 
This two-step formulation is effective by combining the benefits of each individual guidance as discussed in Appendix~\ref{app:visualization}, and provides an interpretable guidance signal.
However, this approach inevitably introduces additional training overhead ($+6\%\text{p}$ per step compared to CFG). A unified, single-step guidance formulation to further improve the guidance-free training efficiency could be a promising future direction.

\paragraph{Extension to visual alignment tasks.}
The SCALE framework currently focuses on improving the sample quality of visual generative models without guidance. However, self-contrastive signals are applicable to broader domains, such as visual alignment tasks. For instance, large visual generative models can generate unsafe, NSFW samples, which can be harmful to society. Self-contrastive signals can be used to pull the models toward the distribution of safe data while pushing it away from unsafe, NSFW distributions, without any guidance at sampling time and without loss of sample quality.

\paragraph{Beyond DPO post-training.}
We empirically verify that SCALE is complementary to DiffusionDPO~\cite{wallace2024diffusion} by post-training a SCALE-trained model with DPO. However, our investigation is currently limited to a preliminary combination with DPO, and a systematic study of how SCALE interacts with RL-based post-training remains an open question. In particular, validating the compatibility of SCALE with more recent RL-based post-training methods, such as DSPO~\cite{zhu2025dspo}, D3PO~\cite{yang2024using}, SPO~\cite{liang2024step}, FlowGRPO~\cite{liu2025flow}, or DanceGRPO~\cite{xue2025dancegrpo}, would be a valuable direction for future work.

\paragraph{Scaling to larger and more recent generative architectures.}
Our experiments cover diffusion models (DiT, SD\,1.5) and AR models (LlamaGen) on academic-scale compute. Recent text-to-image models adopt rectified-flow transformers and a billion-scale parameter regime, including Stable Diffusion~3~\cite{esser2024scaling} and FLUX~\cite{flux2024}. Validating that the self-contrastive signal scales to these flow-based, large-scale models would be a promising direction.

\section{Ethics Statement} \label{app:ethics}

SCALE is a technical framework for improving the visual fidelity of guidance-free generative models. 
%
While SCALE can enhance sample quality and align outputs more closely with learned data distributions without relying on external classifier guidance, the same capabilities could be misused by unauthorized actors. Potential risks include the generation of more convincing, deceptive, or misleading visual media, with downstream harms to privacy, reputation, and public trust. Because SCALE often operates as a post-training method on pretrained backbones while sharing the training datasets, it inherits both the capabilities and the limitations of the underlying models, including any biases, memorization effects, or harmful content patterns present in their training data.

Our experiments are conducted exclusively on publicly available, well-established generative models that ship with safety filters and community-vetted usage policies. SCALE itself is not a safety mechanism; it modifies the sampling trajectory to improve fidelity but does not detect, suppress, or mitigate unsafe content. It should therefore be deployed only in conjunction with robust prompt- and output-level moderation, provenance signals (e.g., watermarking or content credentials) where appropriate, and human oversight for sensitive applications.

This work is intended for academic research and constructive applications such as creative tools, scientific visualization, and the study of guidance mechanisms in diffusion and autoregressive generative models. We explicitly prohibit malicious or unethical use, including the generation of non-consensual imagery, impersonation, disinformation, or any content intended to cause harm to individuals or groups. We encourage practitioners adopting SCALE to carefully document their assumptions and design choices, adhere to the licenses and safety settings of the underlying models, and contribute to the development of clear ethical guidelines, so as to support the responsible advancement of guidance-free generation methods and the broader generative modeling community.

\section{Declaration of LLM Usage} \label{app:llm_usage}

We used LLMs solely for editorial assistance. Their use was limited to polishing grammar, converting handwritten mathematical expressions into \LaTeX{} (including the formatting of proofs in the appendix), and assisting with the visualization of figures and tables (e.g., generating boilerplate plotting code and refining layouts). The model did not contribute to ideation, research direction, or the main claims of this work, and it was not used for data analysis or core algorithmic design. All technical statements, equations, experimental results, and scientific contributions were authored and verified by the authors.


\newpage
\section*{NeurIPS Paper Checklist}

The checklist is designed to encourage best practices for responsible machine learning research, addressing issues of reproducibility, transparency, research ethics, and societal impact. Do not remove the checklist: {\bf The papers not including the checklist will be desk rejected.} The checklist should follow the references and follow the (optional) supplemental material.  The checklist does NOT count towards the page
limit. 

Please read the checklist guidelines carefully for information on how to answer these questions. For each question in the checklist:
\begin{itemize}
    \item You should answer \answerYes{}, \answerNo{}, or \answerNA{}.
    \item \answerNA{} means either that the question is Not Applicable for that particular paper or the relevant information is Not Available.
    \item Please provide a short (1--2 sentence) justification right after your answer (even for \answerNA). 
\end{itemize}

{\bf The checklist answers are an integral part of your paper submission.} They are visible to the reviewers, area chairs, senior area chairs, and ethics reviewers. You will also be asked to include it (after eventual revisions) with the final version of your paper, and its final version will be published with the paper.

The reviewers of your paper will be asked to use the checklist as one of the factors in their evaluation. While \answerYes{} is generally preferable to \answerNo{}, it is perfectly acceptable to answer \answerNo{} provided a proper justification is given (e.g., error bars are not reported because it would be too computationally expensive'' or ``we were unable to find the license for the dataset we used''). In general, answering \answerNo{} or \answerNA{} is not grounds for rejection. While the questions are phrased in a binary way, we acknowledge that the true answer is often more nuanced, so please just use your best judgment and write a justification to elaborate. All supporting evidence can appear either in the main paper or the supplemental material, provided in appendix. If you answer \answerYes{} to a question, in the justification please point to the section(s) where related material for the question can be found.

IMPORTANT, please:
\begin{itemize}
    \item {\bf Delete this instruction block, but keep the section heading ``NeurIPS Paper Checklist"},
    \item  {\bf Keep the checklist subsection headings, questions/answers and guidelines below.}
    \item {\bf Do not modify the questions and only use the provided macros for your answers}.
\end{itemize}


\begin{enumerate}

\item {\bf Claims}
    \item[] Question: Do the main claims made in the abstract and introduction accurately reflect the paper's contributions and scope?
    \item[] Answer: \answerYes{} 
    \item[] Justification: The claim in the abstract and introduction is comprehensively dealt with in the paper, including the background, method, results, and additional analysis in the appendix.

\item {\bf Limitations}
    \item[] Question: Does the paper discuss the limitations of the work performed by the authors?
    \item[] Answer: \answerYes{} 
    \item[] Justification: The limitations are discussed in the appendix, which is also clearly stated in the conclusion of the main manuscript

\item {\bf Theory assumptions and proofs}
    \item[] Question: For each theoretical result, does the paper provide the full set of assumptions and a complete (and correct) proof?
    \item[] Answer: \answerYes{} 
    \item[] Justification: We provide a theoretical justification of the proposed method in the main manuscript and the detailed proof in the appendix.

    \item {\bf Experimental result reproducibility}
    \item[] Question: Does the paper fully disclose all the information needed to reproduce the main experimental results of the paper to the extent that it affects the main claims and/or conclusions of the paper (regardless of whether the code and data are provided or not)?
    \item[] Answer: \answerYes{} 
    \item[] Justification: We provide a detailed experimental setup for reproducing the results in the main manuscript.

\item {\bf Open access to data and code}
    \item[] Question: Does the paper provide open access to the data and code, with sufficient instructions to faithfully reproduce the main experimental results, as described in supplemental material?
    \item[] Answer: \answerNo{} 
    \item[] Justification: We do not include code for the experiments. However, the dataset and baseline code for our implementation is clearly described in the paper, which is fully publicly available.

\item {\bf Experimental setting/details}
    \item[] Question: Does the paper specify all the training and test details (e.g., data splits, hyperparameters, how they were chosen, type of optimizer) necessary to understand the results?
    \item[] Answer: \answerYes{} 
    \item[] Justification: We provide a detailed experimental setup for training and sampling in the main manuscript. The hyperparameters for training is clearly ablated and the effect of hyperparameters for sampling is analyzed in detail. 

\item {\bf Experiment statistical significance}
    \item[] Question: Does the paper report error bars suitably and correctly defined or other appropriate information about the statistical significance of the experiments?
    \item[] Answer: \answerNo{} 
    \item[] Justification: We do not include the error bars for the experiments. However, most of the reported metrics are computed over a large number of samples (50,000 for class-conditional 
    generation and 30,000 for text-to-image generation), following the standard evaluation protocol, which is known to be statistically stable accross different runs.

\item {\bf Experiments compute resources}
    \item[] Question: For each experiment, does the paper provide sufficient information on the computer resources (type of compute workers, memory, time of execution) needed to reproduce the experiments?
    \item[] Answer: \answerYes{} 
    \item[] Justification: In appendix, we report the training environment, as well as required training iterations for each experiment. We also compare the computational analysis of the proposed method with the baseline methods in the main manuscript.
    
\item {\bf Code of ethics}
    \item[] Question: Does the research conducted in the paper conform, in every respect, with the NeurIPS Code of Ethics \url{https://neurips.cc/public/EthicsGuidelines}?
    \item[] Answer: \answerYes{} 
    \item[] Justification: We wrote a paper in compliance with the NeurIPS Code of Ethics.

\item {\bf Broader impacts}
    \item[] Question: Does the paper discuss both potential positive societal impacts and negative societal impacts of the work performed?
    \item[] Answer: \answerYes{} 
    \item[] Justification: We discuss the societal impacts of the proposed method in the appendix.
    
\item {\bf Safeguards}
    \item[] Question: Does the paper describe safeguards that have been put in place for responsible release of data or models that have a high risk for misuse (e.g., pre-trained language models, image generators, or scraped datasets)?
    \item[] Answer: \answerYes{} 
    \item[] Justification: Our work utilizes publicly released large pretrained visual generative model, Stable Diffusion 1.5~\cite{ldm}. We use the model through the official \texttt{diffusers} library, which provides a built-in safety checker that filters NSFW and otherwise unsafe outputs by default. We do not release any large pretrained model that could be potentially misused. We only release the fine-tuning code and model weights that build upon existing checkpoints and publicly available datasets, and inherit the license and safety settings of the underlying models. While a pretrained model is involved (DiT-B/2~\cite{dit}), it poses low risk of misuse as it is relatively small and trained on publicly available, non-sensitive data (ImageNet~\cite{deng2009imagenet}).

\item {\bf Licenses for existing assets}
    \item[] Question: Are the creators or original owners of assets (e.g., code, data, models), used in the paper, properly credited and are the license and terms of use explicitly mentioned and properly respected?
    \item[] Answer: \answerYes{} 
    \item[] Justification: We use open source implementation for our baseline properly denoted in the paper. Otherwise, we do not release any new models pretrained from sensitive, private datasets.

\item {\bf New assets}
    \item[] Question: Are new assets introduced in the paper well documented and is the documentation provided alongside the assets?
    \item[] Answer: \answerNA{} 
    \item[] Justification: We do not release any new assets in the paper.

\item {\bf Crowdsourcing and research with human subjects}
    \item[] Question: For crowdsourcing experiments and research with human subjects, does the paper include the full text of instructions given to participants and screenshots, if applicable, as well as details about compensation (if any)? 
    \item[] Answer: \answerYes{} 
    \item[] Justification: We conduct a blind A/B study with voluntary adult participants. Participation was voluntary and uncompensated, and participants were informed in advance about the purpose of the study and that the displayed images were generated by AI. 

\item {\bf Institutional review board (IRB) approvals or equivalent for research with human subjects}
    \item[] Question: Does the paper describe potential risks incurred by study participants, whether such risks were disclosed to the subjects, and whether Institutional Review Board (IRB) approvals (or an equivalent approval/review based on the requirements of your country or institution) were obtained?
    \item[] Answer: \answerYes{} 
    \item[] Justification: We conduct a blind A/B study with voluntary adult participants. The participants were properly anonymized and detailed description for the study is provided in the appendix.
    The prompts used in the study were sampled from the public dataset, and we manually filtered out any prompts that could be unsafe or sensitive.

\item {\bf Declaration of LLM usage}
    \item[] Question: Does the paper describe the usage of LLMs if it is an important, original, or non-standard component of the core methods in this research? Note that if the LLM is used only for writing, editing, or formatting purposes and does \emph{not} impact the core methodology, scientific rigor, or originality of the research, declaration is not required.
    \item[] Answer: \answerYes{} 
    \item[] Justification: We declare the usage of LLMs for writing, editing, and formatting purposes in the appendix.

\end{enumerate}

\end{document}